\pdfoutput=1

\documentclass[11pt]{article}

\usepackage[final]{acl}
\usepackage{amsmath}  
\usepackage{times}
\usepackage{latexsym}
\usepackage{multirow}

\usepackage{cleveref} 
\usepackage[normalem]{ulem}
\useunder{\uline}{\ul}{}

\usepackage[T1]{fontenc}
\usepackage{booktabs}

\usepackage[utf8]{inputenc}

\usepackage{microtype}
\usepackage{hyperref}

\usepackage{inconsolata}
\usepackage{amsmath}
\usepackage{amsfonts}
\usepackage{graphicx}
\usepackage{tikz}
\usepackage{pgfplots}
\pgfplotsset{compat=1.17}
\usepgfplotslibrary{colormaps}
\pgfplotsset{
colormap={pd}{rgb255(0cm)=(255,255,0) ; rgb255(1cm)=(255,0,0) }
}
\title{StablePrompt: Automatic Prompt Tuning using Reinforcement Learning for Large Language Models}

\author{
 \textbf{Minchan Kwon} \quad
 \textbf{Gaeun Kim} \quad
 \textbf{Jongsuk Kim} \quad
 \textbf{Haeil Lee} \quad
\\
 \textbf{Junmo Kim} \quad
\\
 \textsuperscript{}KAIST
\\
    \texttt{\{kmc0207,
    gkstha5176,
    jskpop,
    haeil.lee,
    junmo.kim\}@kaist.ac.kr}
}

\begin{document}
\maketitle

\begin{abstract}

Finding appropriate prompts for the specific task has become an important issue as the usage of Large Language Models (LLM) has expanded.
Reinforcement Learning (RL) is widely used for prompt tuning, but its inherent instability and environmental dependency make it difficult to use in practice.
In this paper, we propose StablePrompt, which strikes a balance between training stability and search space, mitigating the instability of RL and producing high-performance prompts.
We formulate prompt tuning as an online RL problem between the agent and target LLM and introduce Adaptive Proximal Policy Optimization (APPO).
APPO introduces an LLM anchor model to adaptively adjust the rate of policy updates.
This allows for flexible prompt search while preserving the linguistic ability of the pre-trained LLM.
StablePrompt outperforms previous methods on various tasks including text classification, question answering, and text generation.
Our code can be found in \href{https://github.com/kmc0207/Stableprompt}{github}.
\end{abstract}

\begin{figure*}[t]
    \centerline{\includegraphics[width=\textwidth]{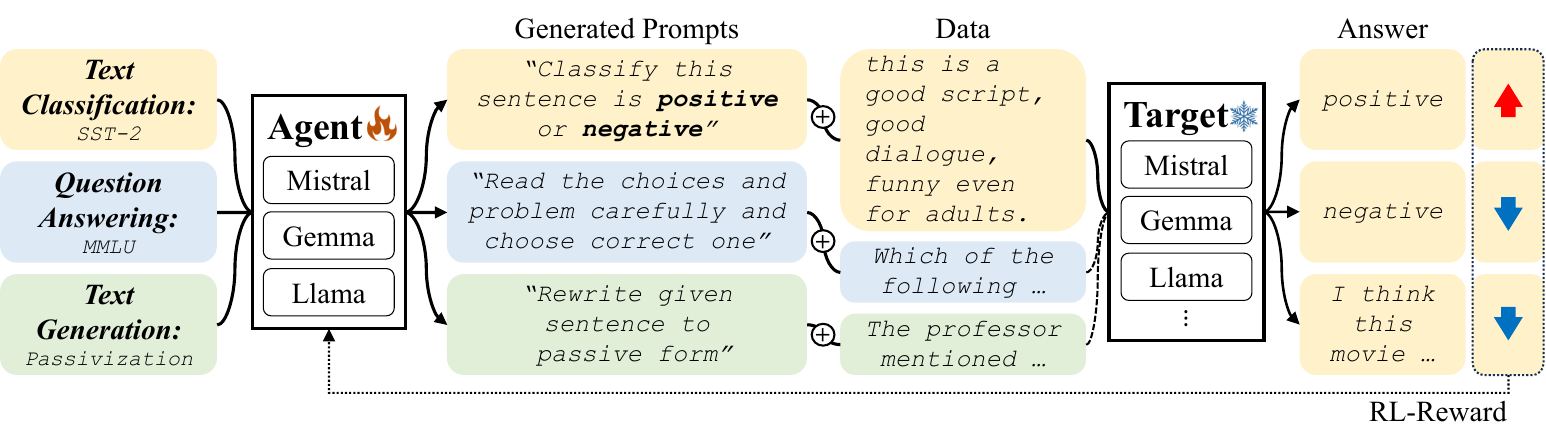}}
    \caption{Overview of StablePrompt. We formulate prompt tuning as an RL-framework using LLMs. We use the target LLM and the given dataset as the world model, and the agent LLM as the policy. We use the response of the target LLM to the prompt generated by the agent LLM as the reward. }
    \label{Fig:Main} 
\end{figure*}

\section{Introduction}
From Semantics~\cite{Breal:1900} to GPT-4~\cite{Achiam:2023}, language models have undergone a huge evolution.
Recently, large language models (LLM) are not only used in traditional natural language processing tasks such as text classification~\cite{wang2018glue} and summarization~\cite{wang2019superglue}, but are also being applied to a wider range of tasks including question answering~\cite{hendrycks2020measuring}, chatting~\cite{ding2023enhancing}, math problem solving~\cite{cobbe2021training}, and planning~\cite{yao2022webshop}.
While LLMs perform well in these areas, they rely heavily on hand-crafted prompts.
Finding or tuning the prompts automatically is crucial to use and evaluate the ability of LLMs in a wider range of applications.
Reinforcement Learning (RL) is a prominent method for prompt tuning due to its ability to update prompts without gradients in the discrete word space.
However, RL is vulnerable to overfitting and is highly dependent on the environment.
This challenge limits the use of RL for a wide variety of LLMs and tasks
Previous methods addressed this by limiting the prompt length~\cite{deng2022rlprompt} or constraining the action space~\cite{zhang2022tempera}, but these approaches reduce the performance of the prompt due to the restricted search space.

In this paper, we propose StablePrompt that keeps training stability while ensuring search space flexibility. 
We define prompt tuning as an online, on-policy RL problem for a given dataset and target LLM.
StablePrompt sets the agent model as the LLM and optimizes the agent model with adaptive proximal policy optimization (APPO).
APPO adaptively adjusts the policy update rate by introducing an anchor model, a snapshot of a point in time on the training trajectory.
This leverages the strong language understanding capabilities of the pre-trained LLMs to give the agent model search space flexibility while maintaining training stability.
We propose two prompt tuning frameworks using APPO: StablePrompt and Test-Time Editing StablePrompt (TTE-StablePrompt). 
StablePrompt generates a single prompt appropriate for the entire dataset, while TTE-StablePrompt generates appropriate prompts for each input.

We validate our methods on a variety of tasks and LLMs. The datasets include text classification~\cite{wang2018glue}, text understanding~\cite{wang2019superglue}, question answering~\cite{hendrycks2020measuring}, and instruction induction~\cite{honovich2022instruction}. 
For agent and target LLMs, we conduct experiments on different sizes ranging from 2B to 13B, using various models such as Llama~\cite{touvron2023llama}, Mistral~\cite{jiang2023mistral}, Gemma~\cite{team2024gemma}, and Falcon~\cite{almazrouei2023falcon}.
To the best of our knowledge, our method is the first RL-based approach that works on agents LLM larger than 7B. 
StablePrompt achieves State-of-The-Art performance across various tasks.

Our contributions are summarized as follows:

\begin{itemize}
    \item  We propose StablePrompt, which is an RL-based prompt tuning method using APPO. APPO introduces an anchor model and modifies the KL-divergence term to keep training stable while ensuring the search space is flexible.
    \item StablePrompt achieves SoTA performance on various tasks including text classification, question answering, and text generation. It can be also used with various types and sizes of agents and target LLMs.
    \item We extend StablePrompt to create an input-dependent prompt. It achieves high performance on tasks that are hard to solve with a single prompt.
\end{itemize}

\section{Related Work}
\subsection{Automatic Prompt Tuning}
Automatic prompt tuning aims to find the appropriate prompts for a given dataset and target model. Soft prompt tuning or Continuous prompt tuning~\cite{bailey2023soft,lester2021power} uses direct gradient descent to search prompts. While it can find the optimal prompt, the generated prompt is often not readable and requires a substantial amount of data to converge. 
By contrast, discrete prompt tuning aims to find prompts in the form of natural language. This approach often operates like black-box optimization, making it suitable for API-based LLMs. Discrete prompt tuning methods can be broadly categorized into generation-based methods and RL-based methods.

\subsection{Discrete Prompt Tuning}

Generation-based methods rely on the text generation abilities of LLMs to find prompts. For example, APE~\cite{zhou2022large} generates prompts by using example input-output pairs, ProTeGi~\cite{pryzant2023automatic} improves prompts through iterative conversation, and PromptAgent~\cite{wang2023promptagent} edits prompts based on a Monte Carlo tree search. Since these methods rely on the performance of a pre-trained LLM without additional tuning, they struggle with tasks that are outside the scope of pre-training.

RL-based methods formulate prompt tuning as an online, on-policy RL problem. For example, GrIPS~\cite{prasad2022grips}, BoostPrompt~\cite{hou2023promptboosting}, and PACE~\cite{dong2023pace} use RL to edit the initial manual prompt. While these methods are relatively stable in training, they heavily depend on the quality of the manual prompt and the predefined action space for editing.
RLprompt~\cite{deng2022rlprompt} is a pioneering work that proposed a method for training agent LLMs using RL. RLprompt adds an MLP layer to the agent LLM for training stability and trains only on this layer. 
However, as the hidden size of the agent LLM increases, the computational overhead increases rapidly.
This is impractical for use in modern LLMs with large hidden sizes.
TEMPERA~\cite{zhang2022tempera} used RL to explore input-dependent prompting. It adopts an agent model that shares a stem of the target LLM to generate input-dependent prompts. 
However, TEMPERA is limited by a predefined action space and struggles with scalability as the hidden size of the target LLM increases.
StablePrompt is designed for a scalable and stable RL-based method.

\begin{figure*}[t]
    \centerline{\includegraphics[width=\textwidth]{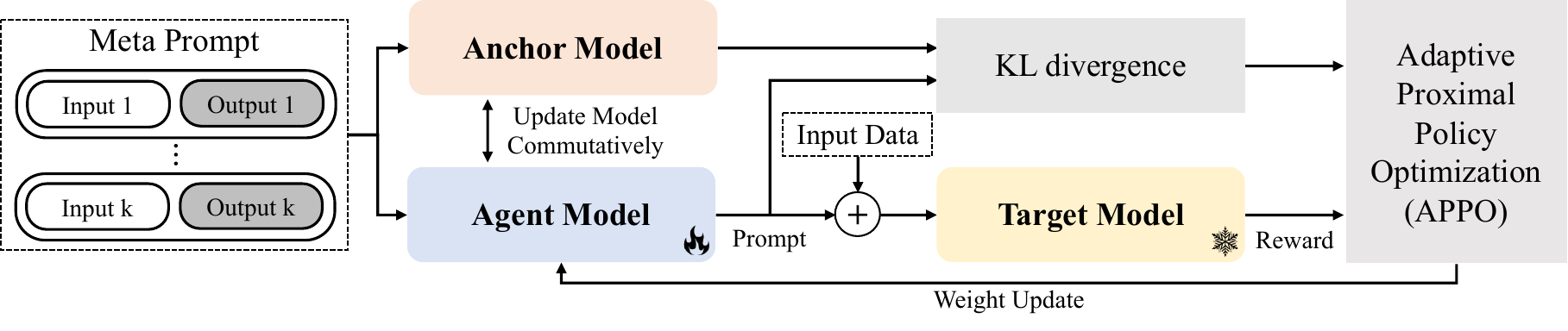}}
    \caption{Training framework of StablePrompt. Generate prompts using the Task agnostic meta-prompt, and calculate the reward of the generated prompts with training data. }
    \label{Fig:Method} 
    \vspace{-1em}
\end{figure*}
\begin{figure}[t]
    \centerline{\includegraphics[width=\columnwidth]{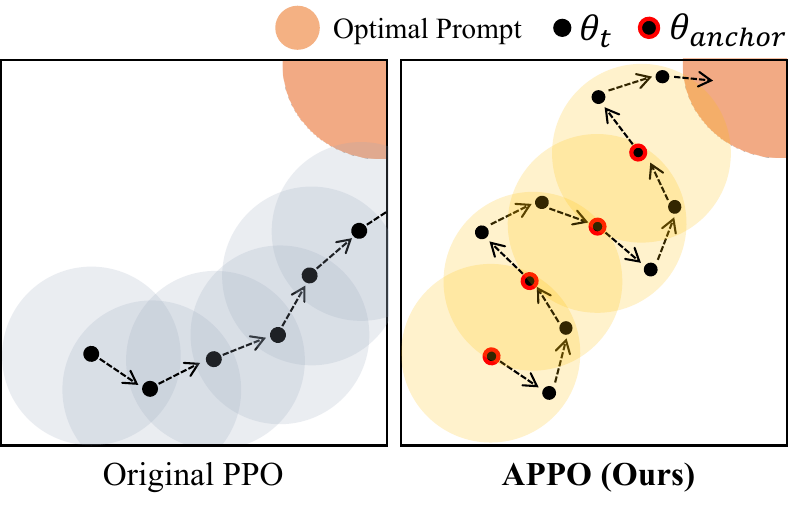}}
    \caption{Illustration comparing APPO to the original PPO. The circle represents the constraint of KL-divergence, and each dot represents the parameter of the agent model at each time step. APPO is robust to incorrect rewards because it maintains an anchor model, while PPO deviates from the optimal prompt as incorrect rewards accumulate.}
    \label{Fig:PPO} 
    \vspace{-1em}
\end{figure}
\section{Method}
\subsection{RL Formulation}
We formulate the discrete prompt tuning as a problem of finding the optimal discrete prompt $\mathbf{z}^\star$ for a given target model $M_T$ and a dataset $D$.
$\mathbf{z}$ is defined in the vocabulary of target model $V$ and prompt length $L$. $\mathbf{z}$ satisfies the following equation.
\begin{equation}
    \operatorname{max}_{\mathbf{z} \in V^L} R(M_T(\mathbf{z},x),y)
\end{equation}
where $R$ is pre-defined reward function, and input-output pair$x,y \in D$.

We introduce an agent model $M_a$ as LLM that generates prompts autoregressively from random input-output pair $(x_r,y_r) \in D$ and task-agnostic meta prompt. We define this set of inputs as state $s$. Detailed meta-prompt can be found in~\Cref{Fig:meta-prompt}. Agent model generates prompts up to the length $l$ according to the $M_a(z_l|s,\mathbf{z}_{<l})$.
After $\mathbf{z}$ is created, it receives a reward from the $R(M_T(\mathbf{z},x),y)$. The full training objective function is below:
\begin{equation} 
    \operatorname{max}_{M_a} R(M_T(\mathbf{z},x),y), \mathbf{z} \sim \displaystyle\prod^L_{l=1} M_a(z_l|s,\mathbf{z}_{<l})
\end{equation}

\paragraph{Original PPO.}
As a method for training LLM agents with RL, we improved Proximal Policy Optimization (PPO).
We refer to the PPO proposed in ~\citet{schulman2017proximal} as the original PPO, to distinguish it from the version modified for RLHF~\citet{ouyang2022training}.
To implement PPO on the LLM agent, we add a value head to the last layer of the LLM agent, which is trained using MSE loss to predict reward values for inputs. 
\begin{equation}
    L_v = (v_{\text{preds}} - \text{reward})^2 
\end{equation}
The value expected from the value head is used with reward to compute advantage $A$, which uses Generalized Advantage Estimation (GAE) and clipped.
\begin{equation}
    A = \operatorname{GAE}(v_{\text{preds}}, \text{reward})
\end{equation}
\begin{equation}
    \text{ratio} = \frac{\theta_{t}(\mathbf{z}|s)}{\theta_{t-1}(\mathbf{z}|s)}
\end{equation}
\begin{equation}
    A_{clipped} = \operatorname{clip}(\text{ratio}, 1-\epsilon,1+\epsilon) * A
\end{equation}
where $\theta$ is parameter of agent model and $t$ is timestep.

Then calculate the penalty $P$ which is the KL-divergence between the previous version of the agent model and the current version. 
The full agent loss is as the follows:
\begin{equation} 
\label{eq : penalty}
    P = \operatorname{KL}(\theta_{t}(\mathbf{z}|s)||\theta_{t-1}(\mathbf{z}|s))
\end{equation}
\begin{equation}
    L_{\text{agent}} = A_{\text{clipped}} + P
\end{equation}

The final PPO objective is as follows:
\begin{equation}
    L_{PPO} = L_v + L_{agent}
\end{equation}

In practice, we perform parameter-efficient training using the LoRA~\cite{hu2021lora} and update only the value head and the LoRA adaptor.

\subsection{StablePrompt}

\paragraph{Anchor Model.}
Traditional PPO methods limit updates relative to the previous step, making it difficult to prevent errors from accumulating.
We introduce an anchor model, which is a copy of the agent model with validated performance improvements in the training trajectory. 
The anchor model starts as a copy of the initial agent and is carefully updated at a predefined update period $u_t$. 
If the performance of the current agent model is higher than an update threshold compared to the anchor model, the anchor model is updated to the copy of the current agent model. 
Conversely, if the agent model underperforms the anchor model by less than a rollback threshold, the agent model is rolled back to the anchor model.

This allows the anchor model to adaptively update based on the characteristics of the task. If the reward signal is stable or requires several update steps to find the optimal prompt, the anchor model is updated accordingly. On the other hand, if the reward signal is unstable or does not require many updates to find the optimal prompt, the anchor model is updated in a few steps or not. 

\paragraph{Adaptive PPO.}
 The KL-divergence penalty term (\Cref{eq : penalty}) uses the parameters of the previous model to prevent the current model from changing too much. But as the steps get longer, the model can gradually diverge from the initial. When unstable reward signals accumulate, this can lead the model into a local minima.

In RLHF-style PPO~\cite{ouyang2022training}, the penalty term (\Cref{eq : penalty}) is modified by $\operatorname{KL}(\theta_{t}(\mathbf{z}|s)||\theta_{0}(\mathbf{z}|s))$ to prevent the agent model from deviating too far from the initial version. 
This is appropriate for a task like RLHF that needs to answer a wide variety of questions while not losing the initial language generation capability.
However, in prompt tuning, RLHF-style PPO limits the search space of the agent to the initial agent, which leads to suboptimal prompts.

We propose Adaptive PPO (APPO), which combines the advantages of RLHF-style and original PPO, achieving both training stability and an extensive search space.
We leverage the anchor model to modify \Cref{eq : penalty} as follows :
\begin{equation}
    P_{\text{APPO}} = KL(\theta_{t}(\mathbf{z}|s)||\theta_{\text{anchor}}(\mathbf{z}|s))
\end{equation}

This term restricts the agent model from diverging too far from an anchor model. This approach allows for more conservative agent updates compared to the original PPO while ensuring a larger search space compared to RLHF-style PPO.
The full objective of APPO is below:
\begin{equation}
    L^{\text{APPO}}_{\text{agent}} = A_{\text{clipped}} + P_{\text{APPO}}
\end{equation}
\begin{equation}
    L_{\text{APPO}} = L_v + L^{\text{APPO}}_{\text{agent}}
\end{equation}

\paragraph{Reward Function.}
We design reward functions for two tasks: text classification and text generation. For text classification, we use accuracy and softmax difference. While accuracy is a good reward function, it has discrete values, which can lead to many prompts having the same accuracy. This problem is often encountered in scenarios with limited training data, such as few-shot text classification. To mitigate this, we introduce the softmax difference, which subtracts the highest value among the incorrect options from the value of the correct answer from the softmax output. 
The following expression combines accuracy and the softmax difference, with coefficients $c_a$ and $c_s$ applied respectively.
\begin{equation}
    R(\mathbf{z},x,y) = c_a \operatorname{Acc}(\mathbf{z},x,y) + c_s \operatorname{D}(\mathbf{z},x,y)
\end{equation}
\begin{equation}
    \operatorname{D} = M_T(\mathbf{z},x)_{i=y} - \operatorname{max}[M_T(\mathbf{z},x)_{i\neq y}]
\end{equation}
This metric is used to rank prompts when they have the same accuracy. The softmax difference is also widely used in other RL frameworks for classification~\cite{han2023reinforcement}.

For text generation, we use the F1 score directly as the reward function. 

\begin{table*}[]
\resizebox{\textwidth}{!}{%
\begin{tabular}{@{}ccccccccc@{}}
\toprule
                                        & Method            & SST-2       & MRPC      & RTE       & QNLI      & MNLI      & SNLI & Average \\ \midrule
\multirow{2}{*}{Fine-Tuning}            & Fine-Tuning       & 71.9        & 59.6      & 55.7      & 63.1      & 41.1      &64.8  & 59.3     \\
                                        & Soft prompt tuning& 78.3 & 57.1 &51.6 & \textbf{89.0} & 34.9 &
                                        55.8  & 61.1 \\ \midrule
\multirow{3}{*}{Fixed prompt}           & Manual prompt & {\ul 89.1} & 51.0 &
                                        64.0  & 73.0  & {\ul 67.0}  &
                                        47.0  & 65.2  \\
                                        & Zero-shot CoT & 57.9  & 38.4  &
                                        \textbf{81.6}  & 75.2 & \textbf{71.1}  &
                                        66.3  & 65.1  \\
                                        & Few-shot prompt & 55.0  & 49.0  &
                                        76.0  & { \ul 82.0}  & 58.0  &
                                        52.2  & 62.0  \\ \midrule
\multirow{7}{*}{Discret prompt tuning}  & GrIPS & 84.7($\pm$4.6)	&55.6($\pm$2.6)&	60.9($\pm$3.5)&	28.9($\pm$1.2)	&44.4($\pm$1.1)&	63.5($\pm$2.3) & 59.4 \\
                                        & PromptBoosting &65.4($\pm$1.0)	&52.7($\pm$1.1)&	71.6($\pm$0.9)&	71.6($\pm$1.1)&	35.5($\pm$1.4)&	52.6($\pm$1.8) & 58.2\\
                                        & APE &83.2($\pm$7.7)	&55.3($\pm$4.9)	&78.6($\pm$1.3)	&75.0($\pm$2.2)	&54.6($\pm$7.9)	& {\ul 72.3 } ($\pm$4.8) & {\ul 70.1} \\
                                        & ProTeGi &69.2($\pm$8.4)	&48.8($\pm$1.3)&	73.2($\pm$6.3)	&74.2($\pm$7.7)	&56.6($\pm$10.9)	&61.3($\pm$12.3) &64.0\\
                                        & RLprompt &70.8($\pm$6.5)	&{\ul 56.0} ($\pm$1.5)	&67.3($\pm$2.5)	&62.6($\pm$1.3)	&54.6($\pm$1.9)	&56.6($\pm$1.3) &61.3\\
 \cline{2-9} 
                                        & \begin{tabular}[c]{@{}c@{}}StablePrompt\\ (Ours)\end{tabular} &\textbf{92.5}($\pm$1.3)&	\textbf{71.3}($\pm$3.4)&	{\ul 81.5}($\pm$2.8)&	 75.9 ($\pm$1.4)	&63.3($\pm$1.2)&	\textbf{74.1}($\pm$1.4)& \textbf{76.4} \\ \bottomrule
\end{tabular}%
}
\caption{Result for 6 few-shot text classification datasets. StablePrompt outperforms other discrete prompt tuning methods. Generated prompts can found in~\Cref{Generated_Prompt_FSTC}}
\label{tab:TC}
\end{table*}


\subsection{Test Time Editing StablePrompt.}
For tasks that are difficult to solve with a single prompt, we expand StablePrompt to generate prompts that depend on the input query. We call this extended version as Test-Time Editing StablePrompts (TTE-StablePrompt).

In TTE-StablePrompt, the input state $s$ consists of a meta prompt that includes randomly chosen input-output pairs and the current input.
The agent generates a prompt for the current input. The generated prompt and current input are fed into the target model to calculate rewards. Detailed meta prompt can be found in~\Cref{Fig:meta-prompt}. We keep the same settings for the other parts of the method.

This approach is different from StablePrompt, which uses the average value of the training batch as its reward.
The reward of TTE-StablePrompt is calculated using only the current input. 
In TTE-StablePrompt, the instance reward signals train the agent model to generate prompts optimized for specific inputs, rather than the entire dataset.


\begin{figure}[t]
\centering
\begin{tikzpicture}
\begin{axis}[
    colormap name=pd, 
    point meta min=47.7,
    point meta max=95,
    ytick=data,
    yticklabels={G2,G7,M7,L8,F11},
    xtick=data,
    xticklabels={MP,G2,G7,M7,L8},
    xtick pos=top,
    enlarge x limits=false,
    enlarge y limits=false,
    ylabel=Target Model,
    xlabel=Agent Model,
    width=0.75\columnwidth, 
    height=0.75\columnwidth, 
    mesh/cols=5, 
    nodes near coords,
    every node near coord/.append style={font=\small, inner sep=0pt}
]
\addplot [matrix plot*,point meta=explicit] table [meta=val] {
    x y val
    0 0 51.0
    1 0 62.5
    2 0 63.1
    3 0 62.5
    4 0 61.7
    0 1 71.2
    1 1 73.6
    2 1 76.0
    3 1 75.4
    4 1 74.9
    0 2 62.5
    1 2 77.5
    2 2 78.9
    3 2 79.3
    4 2 79.8
    0 3 55.4
    1 3 54.9
    2 3 59.5
    3 3 62.7
    4 3 64.9
    0 4 70.9
    1 4 74.8
    2 4 78.8
    3 4 78.7
    4 4 78.2
};
\end{axis}
\end{tikzpicture}
\caption{Heatmap of few-shot text classification tasks on diverse target-agent pairs. Reported numbers are an average of 6 datasets. \textit{MP : Manual prompt, G2: Gemma-2B, G7: Gemma-7B, M7: Mistral-7B,
L8: Llama-3-8B, F11: Falcon-11B.} 
StablePrompt works well with a variety of LLMs.}
\label{fig:agent-target pair}
\end{figure}
\section{Experiment}
\subsection{Few Shot Text Classification} \label{few-shot}
\paragraph{Datasets.}
Few-shot text classification is used in many previous prompt tuning studies, including ~\citet{deng2022rlprompt,zhang2022tempera}. 
We use the subsets of GLUE~\cite{wang2018glue} and SuperGLUE~\cite{wang2019superglue}, including sentiment analysis datasets (SST-2) and natural language inference datasets (MRPC, MNLI, QNLI, SNLI, and RTE). 

For inference, we use a verbalizer with predefined class label tokens. When determining the prediction of models, we select candidates only from the set of verbalizers. Detailed dataset statistics and verbalizer settings can be found in the~\Cref{tab:appendix_fewshot-dataset-detail}.

\paragraph{Baselines.}
Our baselines include supervised fine-tuning methods such as LoRA fine-tuning and soft prompt tuning~\cite{bailey2023soft}. We also use fixed prompts including hand-crafted manual prompts, few-shot prompts, and zero-shot chain of thought (Zero-Shot CoT) prompts~\cite{wei2022chain}.

For direct comparison with StablePrompt, we use various discrete prompt tuning methods. These include generation-based methods such as APE~\cite{zhou2022large} and ProTeGi~\cite{pryzant2023automatic}, and RL-based method such as GrIPS~\cite{prasad2022grips}, PromptBoosting~\cite{hou2023promptboosting} 
 and RLprompt~\cite{deng2022rlprompt}, which is directly comparable to ours.
 Therefore, we conduct experiments on the 330M RoBERTa-large ~\cite{liu2019roberta}  model and include the results in the~\Cref{tab:roberta-fewshot}.

\begin{table*}[!t]
\resizebox{\textwidth}{!}{%
\begin{tabular}{ccccccc}
\toprule
Method & Human prompt & \begin{tabular}[c]{@{}c@{}}Human prompt \\ + PACE\end{tabular} & \multicolumn{3}{c}{APE} & StablePrompt \\ \midrule
Agetn Model              & -    & -    & GLM            & OPT             & InstructGPT3.5  & Mistral \\
Parameters          & -    & -    & 130B           & 175B            & unknown         & 7B                 \\ \hline
Antonyms            & {\ul 85.0} & \textbf{87.0} & 78.0($\pm$0.5) & 82.7($\pm$0.7)  & 81.0($\pm$0.7)  & 83.7($\pm$0.9)     \\
Cause selection     & 84.0 & {\ul 85.0} & 53.3($\pm$0.1) & 65.3($\pm$1.0)  & 72.0($\pm$1.0)  & \textbf{88.7}($\pm$1.0)     \\
Passivization       & 100.0 & 100.0 & 7.3($\pm$0.0)  & 100.0($\pm$0.0) & 100.0($\pm$0.0) & 100.0($\pm$0.0)    \\
Second Letter       & 99.0 & 100.0 & 3.3($\pm$0.9)  & 100.0($\pm$0.0) & 100.0($\pm$0.0) & 100.0($\pm$0.0)    \\
Sentiment           & {\ul 91.0} & \textbf{92.0} & 87.7($\pm$0.8) & 82.7($\pm$0.9)  & 88.3($\pm$0.8)  & 90.7($\pm$0.9)     \\
Translation en-fr   & {\ul 89.0 } & 88.0 & 79.7($\pm$0.8) & 85.3($\pm$0.8)  & 84.3($\pm$0.8)  & \textbf{90.3}($\pm$1.0)     \\ \hline
Average on 6 tasks             & 91.3 & {\ul 92.0} & 51.8           & 68.6            & 89.3            & \textbf{92.8}               \\
Average on 24 tasks & 79.8 & {\ul 80.3} & -              & -               & 77.5            & \textbf{81.5}               \\ \bottomrule
\end{tabular}%
}
\caption{Result for 6 selected tasks and an average of all 24 tasks in the Instruction induction dataset with InstructGPT3.5 as the target model. Full results can be found in \Cref{tab:appendix_gpt_full}.}
\label{tab:GPT}
\end{table*}
\begin{table}[]
\resizebox{\columnwidth}{!}{%
\begin{tabular}{@{}cccc@{}}
\toprule
              & \multicolumn{2}{c}{BBII} & II          \\
 &
  \begin{tabular}[c]{@{}c@{}}Text \\ Classification\end{tabular} &
  \begin{tabular}[c]{@{}c@{}}Text \\ Generation\end{tabular} &
  \begin{tabular}[c]{@{}c@{}}Instruction\\ Induction\end{tabular} \\ \midrule
Manual Prompt & 51.57        & 37.61       & 33.70       \\
PromptAgent   & 28.50        & -           & -           \\
APE           & 56.46        & 49.59       & {\ul 51.94}       \\
ProTeGi       & {\ul 56.58}  & {\ul 55.61}       & 51.60       \\

\midrule
\begin{tabular}[c]{@{}c@{}}StablePrompt\\ (Ours)\end{tabular} &
  \textbf{57.75} &
  \textbf{61.36} &
  \textbf{65.80} \\ \bottomrule
\end{tabular}%
}
\caption{Result for BigBench-Hard Instruction Induction (BBII) and Instruction Induction (II) datasets. For BBII, we divided it into two parts based on the type of task. Full results can be found in ~\Cref{tab:appendix_BBH_full} and~\Cref{tab:appendix_ii_full}.}
\label{tab:BBII}
\end{table}
\paragraph{Implementation Details.}
We perform two settings on Few-Shot Text Classification.
One is an experiment with both the target and the agent model fixed to gemma-1.1-7B-it (Gemma-7B) ~\cite{team2024gemma} for comparison with the baselines. 
For RLprompt, we use GPT2-XL~\cite{radford2019language} as the agent due to computational overhead.

The other experiment runs StablePrompt on five target models: gemma-1.1-2B-it (Gemma-2B), Gemma-7B, Mistral-7B-it-v2.0 (Mistral-7B)~\cite{jiang2023mistral}, llama3-8B-it (Llama3-8B)~\cite{touvron2023llama}, and falcon-11B (Falcon-11B)~\cite{almazrouei2023falcon}, and four agent models: Gemma-2B, Gemma-7B, Mistral-7B, Llama3-8B. We report the average accuracy of 6 datasets in~\Cref{tab:TC}.

All experiments are performed with three distinct random seeds.
For the generated prompts, we use the template \textit{"[prompt] Input : [input] Output :"} for prediction. 
We use only 16 samples per label for training. 
The generated prompts of each step are queued in pairs with rewards. At the test time, the top 5 prompts in order of reward are selected for testing and report the highest performance. This is the same method as RLprompt.
Detailed numbers are shown in the~\Cref{tab:RL_parameter}.

\paragraph{Results.}
~\Cref{tab:TC} shows the performance of various baselines and StablePrompt. StablePrompt achieves State-of-The-Art (SoTA) performance on all tasks except QNLI. 
In QNLI, StablePrompt also achieves the best performance among the discrete prompt tuning methods.
The average score also outperforms APE and achieves SoTA.
We present the full generated prompt in the~\Cref{Generated_Prompt_FSTC}.

~\Cref{fig:agent-target pair} illustrates the performance of StablePrompt across various Agent-Target pairs. The values in the heatmap are the averages of six datasets. StablePrompt outperforms manual prompts across all pairs except (Gemma-2B, Llama3-8B) pair. These results demonstrate that our method is robust to model sizes, such as a small agent model of 2B and a large target model of 11B. 

Specifically, when comparing Mistral-7B and Falcon-11B, the manual prompt performance is higher with falcon-11B, but with appropriate prompting from StablePrompt, Mistral-7B outperforms falcon-11B. This demonstrates that an appropriate prompt allows a small model to easily understand a task and achieve performance comparable to a large model.

\begin{table*}[]
\centering
\resizebox{0.9\textwidth}{!}{%
\begin{tabular}{@{}ccccccc@{}}
\toprule
 Datasets& \multicolumn{5}{c}{MMLU} & OpenbookQA \\ 
 \cmidrule(lr){1-1}  \cmidrule(lr){2-6}   \cmidrule(lr){7-7} 
 Subsets & STEM & \begin{tabular}[c]{@{}c@{}}Social \\ Sciences\end{tabular} & Humanities & Other & Average & \begin{tabular}[c]{@{}c@{}}Average\end{tabular} \\ \midrule
manual prompt + fewshot & 47.1 & 61.6 & 55.4 & 54.5 & 53.9 & 62.6 \\
Zero-Shot CoT & {\ul 49.2} & 59.6 & 54.5 & 56.0 & 54.2 & - \\
APE & 45.0 & 59.3 & 56.4 & 51.1 & 52.1 & 70.7 \\
ProTeGi & 45.7 & 59.7 & 56.0 & 55.3 & 53.3 & 71.5 \\
RLprompt & 46.5 & 55.1 & 56.6 & 55.7 & 52.8 & 63.6 \\ \midrule
StablePrompt (Ours) & 47.8 & {\ul 63.6} & {\ul 58.6} & {\ul 59.0} & {\ul 56.3} & {\ul 72.2} \\
TTE-StablePrompt (Ours) & \textbf{49.6} & \textbf{65.7} & \textbf{59.6} & \textbf{58.8} & \textbf{57.5} & \textbf{78.7} \\ \bottomrule
\end{tabular}%
}
\caption{Results for QA tasks. We use MMLU and OpenbookQA datasets with Gemma-7B as the target model. Full results can be found in~\Cref{tab:appendix_mmlu_full}.}
\label{tab:MMLU}
\end{table*}

\subsection{Induction Task}
\paragraph{Datasets.}
We experiment with an induction task in which the agent has to provide a rule for an input-output pair as a prompt.
We use the Instruction Induction dataset (II)~\cite{naturalinstructions} and BigBench-Instruction Induction dataset (BBII)~\cite{zhou2022large}, a subset of BiG-Bench~\cite{ghazal2013bigbench}. These include tasks such as editing the input sentence or finding answers according to rules. Each task requires prompts in the form of instructions designed to help the target model induce the correct answer.

The tasks consist of text classification and text generation, requiring an understanding of various fields such as spelling, morphosyntax, and phonetics. We conduct experiments on BBII, which has 20 subsets, and Instruction Induction, which has 23 subsets. The dataset details can be found in~\Cref{appendix_detail_Induction}.

\paragraph{Implementation Details.}
We perform experiments with two different target models.
One is the Gemma-7B and the other is InstructGPT3.5.
For the first experiments, due to the large number of datasets, we use APE and ProTeGi as baselines, and we include PromptAgent~\cite{wang2023promptagent} which is a Monte Carlo tree search-based generation method designed for BigBench text classification tasks.

In experiments with InstructGPT3.5, we use APE to reduce the number of steps due to cost. For APE, we use various agent models such as ~\cite{zhang2022opt,zeng2022glm}.
We use PACE~\cite{dong2023pace}, an RL-based editing method designed for induction tasks, and a human prompt from the same paper as the baseline.

For text classification, we use the same reward function as~\Cref{few-shot}. For text generation, we use the F1 score as a reward function. We use the same template as~\Cref{few-shot} for both BBII and II.

\paragraph{Results.}
Experiments on the Gemma-7B target model are presented in~\Cref{tab:BBII}. 
Our method achieves SoTA on both BBII and II. 
In particular, it outperforms the text generation tasks II and BBII.
This shows the effectiveness of the RL framework on the text generation tasks, where the format of the output is important.

~\Cref{tab:GPT} shows the experiments conducted using InstructGPT3.5. StablePrompt outperforms even when using the large black-box model InstructGPT3.5 as the target model.
This highlights the benefits of the RL-based method, which works well when the target model is not publicly accessible.

Note that our method outperforms APE, which uses models larger than 100B as the agent. 
In particular, the 7B model trained by StablePrompt produces better prompts than the commercial black box model InstructGPT3.5. 
This shows that our method does not rely on the ability of the agent model and is cost-efficient by using a small model.

\subsection{Question Answering}
\paragraph{Datasets.}
We conduct an experiments on a Question Answering (QA) task.
In this paper, we use the MMLU~\cite{hendrycks2020measuring} and OpenbookQA~\cite{OpenBookQA2018} dataset, which requires users to answer questions from various fields. We report the performance of 57 question topics from MMLU, categorized into STEM, Humanity, Social Science, and Others. For OpenbookQA, in addition to the question, a fact relevant to each question is provided as a hint, which we include as a prompt before the question during experiments. The verbalizer is used in the same way as for text classification. We present 4 options (A,B,C,D) in a question and use the alphabet corresponding to each option as a verbalizer. The reward function is the same as~\Cref{few-shot}. Detailed numbers of datasets can be found in~\Cref{appendix_detail_qa}.

\paragraph{Implementation Details.}
The target and agent models are both fixed with Gemma-7B. For the prompt, we use the template \textit{"[Prompt] Question : [Question] Choice : [Choice] Output :"}. We train the model using 20 question-answer pairs from the validation dataset for each topic.

\begin{figure*}[t]
    \centerline{\includegraphics[width=\textwidth]{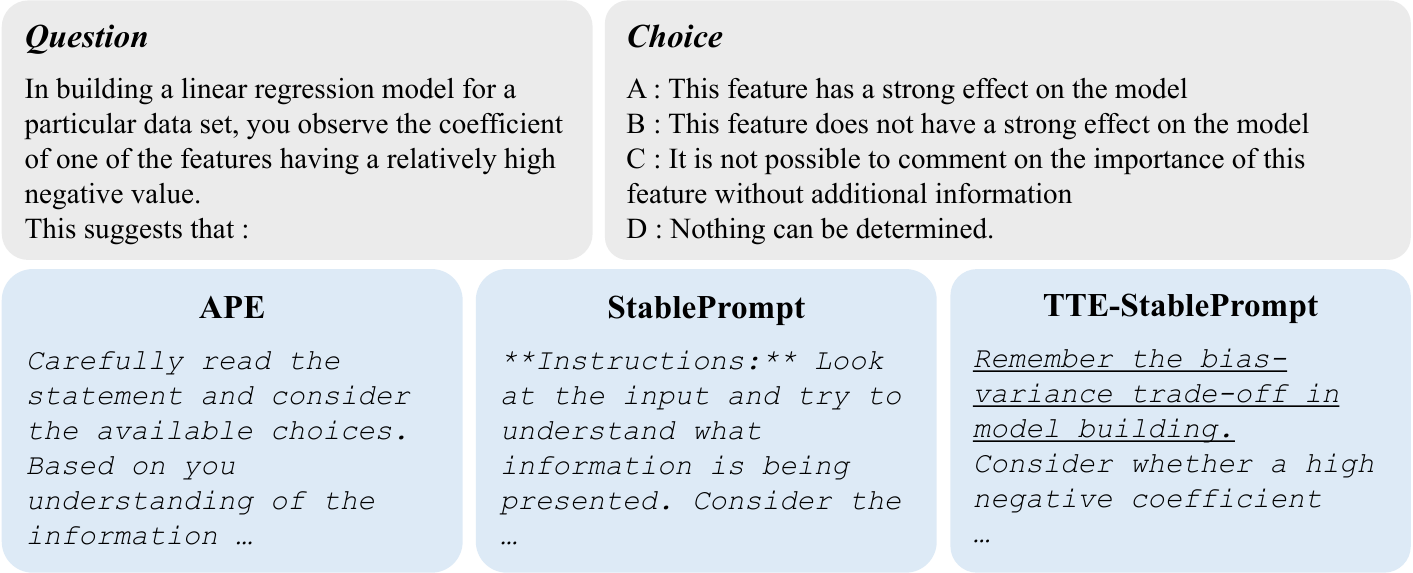}}
    \caption{Generated prompts and input in machine learning subset of MMLU dataset. We truncate the latter part of the generated prompt for readability. The full prompt can be found in~\Cref{appendix_figure5}}
    \label{Fig:QA_TTA} 
\end{figure*}
\begin{table}[]
\centering
\resizebox{\columnwidth}{!}{%
\begin{tabular}{@{}lccc@{}}
\toprule
 &
  \begin{tabular}[c]{@{}c@{}}Original \\PPO\end{tabular} &
  \begin{tabular}[c]{@{}c@{}}RLHF-style \\PPO\end{tabular} &
  \begin{tabular}[c]{@{}c@{}}APPO\\ (Ours)\end{tabular} \\ \midrule
SST2    & {\ul 91.5}($\pm$0.7)          &  91.1($\pm$1.0)    & \textbf{92.5}($\pm$1.3) \\
MRPC    & 65.9($\pm$9.0)          & {\ul 70.6}($\pm$2.2) & \textbf{71.3}($\pm$3.4)    \\
RTE     & 80.2($\pm$2.1)          & {\ul 80.3}($\pm$2.0)    & \textbf{81.5}($\pm$2.8) \\
QNLI    & 70.2($\pm$2.1)          & \textbf{76.7}($\pm$1.6) & {\ul 75.9}($\pm$1.4)    \\
MNLI    & \textbf{66.2}($\pm$2.5) & 61.0($\pm$1.2)          & {\ul 63.3}($\pm$1.2)    \\
SNLI    & 69.5($\pm$1.9)          & {\ul 70.4}($\pm$3.3)          & \textbf{74.1}($\pm$1.4) \\ \midrule
Average & 73.3                    & {\ul 74.2}              & \textbf{76.4}           \\ \bottomrule
\end{tabular}%
}
\caption{Result for ablation study of PPO components on few-shot text classification tasks. We report the average and standard deviation of experiments from 5 distinct random seeds.}
\label{tab:ablation}
\end{table}

\paragraph{Results.}
~\Cref{tab:MMLU} shows the performance of various baselines. StablePrompt achieves the highest performance among the baselines.
In particular, StablePrompt shows comparable results with STEM while outperforming all other methods.

There are many different questions on the same topic that are difficult to solve with a single prompt. 
TTE-StablePrompt, which gives different instructions depending on the input within the same subject, is more effective than StablePrompt, which only uses a single prompt. TTE-StablePrompt also performs better than Zero-Shot CoT, which uses the same multi-step reasoning and is known to perform well on maths and science tasks.

~\Cref{Fig:QA_TTA} shows question-choice pairs from the machine learning dataset in MMLU, along with the prompts generated by APE, StablePrompt, and TTE-StablePrompt. APE and StablePrompt generate almost semantically similar prompts which can be generally used for all questions in subject.
However, TTE-StablePrompt generates prompts appropriate to the given question (emphasized with underlining). This shows that a simple TTE extension effectively creates an input-dependent prompt.

\subsection{Ablation Study}
\paragraph{Experiment Settings.}
We conduct an ablation study for APPO. We use the same settings as few-shot text classification. We fix the agent and the target model to Gemma-7B. 

\paragraph{Results.}
~\Cref{tab:ablation} shows the performance of PPO variants.
APPO outperforms on average across all tasks by leveraging the strengths of both the original PPO and RLHF-style PPO through adaptive anchor model updates. APPO can either behave like the RLHF-style PPO, with no updates, or like the original PPO, with updates in each update period. In particular, when the performance gap between the original PPO and RLHF-style PPO is significant, APPO adapts to the better-performing model. This pattern is observed in tasks like MRPC, QNLI, and MNLI. Additionally, in tasks such as SNLI, APPO can identify more appropriate prompts than either the original or RLHF-style PPO alone. This shows that APPO takes advantage of PPO and RLHF-style PPO and reaches a better convergence point.

\section{Conclusion}
In this paper, we propose a novel RL-based prompt tuning method, StablePrompt. 
StablePrompt defines prompt tuning as an online, on-policy RL problem and introduces APPO.
We validate that StablePrompt outperforms than other methods across various target models and tasks. 
To the best of our knowledge, this is the first RL-based prompt tuning method for models larger than 7B. 
StablePrompt demonstrates the potential to integrate existing RL methodologies into prompt tuning, and we believe there is a capacity for further expansion of RL-based prompt tuning approaches.

\section*{Limitation} 
The limitations of this study can be summarized as follows: 
(1) This paper does not cover experiments that are significantly beyond the scope of prior learning, such as medical and legal domains; however, since it is a training-based method, it is expected to be scalable in future work.
(2) This paper can be used to abuse LLM for specific purposes. This is a particular threat to commercial LLMs in the API format because they are based on black-box optimization.

\section*{Acknowledgments}
This work was supported by Institute of Information \& Communications Technology Planning \& Evaluation(IITP) grant funded by the Korea government(MSIT) (RS-2024-00439020, Developing Sustainable, Real-Time Generative AI for Multimodal Interaction, SW Starlab)

\bibliography{custom}

\begin{thebibliography}{35}
\providecommand{\natexlab}[1]{#1}

\bibitem[{Achiam et~al.(2023)Achiam, Adler, Agarwal, Ahmad, Akkaya, Aleman, Almeida, Altenschmidt, Altman, Anadkat et~al.}]{Achiam:2023}
Josh Achiam, Steven Adler, Sandhini Agarwal, Lama Ahmad, Ilge Akkaya, Florencia~Leoni Aleman, Diogo Almeida, Janko Altenschmidt, Sam Altman, Shyamal Anadkat, et~al. 2023.
\newblock Gpt-4 technical report.
\newblock \emph{arXiv preprint arXiv:2303.08774}.

\bibitem[{Almazrouei et~al.(2023)Almazrouei, Alobeidli, Alshamsi, Cappelli, Cojocaru, Debbah, Goffinet, Heslow, Launay, Malartic et~al.}]{almazrouei2023falcon}
Ebtesam Almazrouei, Hamza Alobeidli, Abdulaziz Alshamsi, Alessandro Cappelli, Ruxandra Cojocaru, Merouane Debbah, Etienne Goffinet, Daniel Heslow, Julien Launay, Quentin Malartic, et~al. 2023.
\newblock Falcon-40b: an open large language model with state-of-the-art performance.
\newblock \emph{Findings of the Association for Computational Linguistics: ACL}, 2023:10755--10773.

\bibitem[{Bailey et~al.(2023)Bailey, Ahdritz, Kleiman, Swaroop, Doshi-Velez, and Pan}]{bailey2023soft}
Luke Bailey, Gustaf Ahdritz, Anat Kleiman, Siddharth Swaroop, Finale Doshi-Velez, and Weiwei Pan. 2023.
\newblock Soft prompting might be a bug, not a feature.

\bibitem[{Br{\'e}al(1900)}]{Breal:1900}
Michel Br{\'e}al. 1900.
\newblock \emph{Semantics: Studies in the science of meaning}.
\newblock W. Heinemann.

\bibitem[{Cobbe et~al.(2021)Cobbe, Kosaraju, Bavarian, Chen, Jun, Kaiser, Plappert, Tworek, Hilton, Nakano et~al.}]{cobbe2021training}
Karl Cobbe, Vineet Kosaraju, Mohammad Bavarian, Mark Chen, Heewoo Jun, Lukasz Kaiser, Matthias Plappert, Jerry Tworek, Jacob Hilton, Reiichiro Nakano, et~al. 2021.
\newblock Training verifiers to solve math word problems.
\newblock \emph{arXiv preprint arXiv:2110.14168}.

\bibitem[{Deng et~al.(2022)Deng, Wang, Hsieh, Wang, Guo, Shu, Song, Xing, and Hu}]{deng2022rlprompt}
Mingkai Deng, Jianyu Wang, Cheng-Ping Hsieh, Yihan Wang, Han Guo, Tianmin Shu, Meng Song, Eric~P Xing, and Zhiting Hu. 2022.
\newblock Rlprompt: Optimizing discrete text prompts with reinforcement learning.
\newblock \emph{arXiv preprint arXiv:2205.12548}.

\bibitem[{Ding et~al.(2023)Ding, Chen, Xu, Qin, Zheng, Hu, Liu, Sun, and Zhou}]{ding2023enhancing}
Ning Ding, Yulin Chen, Bokai Xu, Yujia Qin, Zhi Zheng, Shengding Hu, Zhiyuan Liu, Maosong Sun, and Bowen Zhou. 2023.
\newblock Enhancing chat language models by scaling high-quality instructional conversations.
\newblock \emph{arXiv preprint arXiv:2305.14233}.

\bibitem[{Dong et~al.(2023)Dong, Luo, Jiang, Jin, and Li}]{dong2023pace}
Yihong Dong, Kangcheng Luo, Xue Jiang, Zhi Jin, and Ge~Li. 2023.
\newblock Pace: Improving prompt with actor-critic editing for large language model.
\newblock \emph{arXiv preprint arXiv:2308.10088}.

\bibitem[{Ghazal et~al.(2013)Ghazal, Rabl, Hu, Raab, Poess, Crolotte, and Jacobsen}]{ghazal2013bigbench}
Ahmad Ghazal, Tilmann Rabl, Minqing Hu, Francois Raab, Meikel Poess, Alain Crolotte, and Hans-Arno Jacobsen. 2013.
\newblock Bigbench: Towards an industry standard benchmark for big data analytics.
\newblock In \emph{Proceedings of the 2013 ACM SIGMOD international conference on Management of data}, pages 1197--1208.

\bibitem[{Han et~al.(2023)Han, Choi, Lee, and Kim}]{han2023reinforcement}
Gyojin Han, Jaehyun Choi, Haeil Lee, and Junmo Kim. 2023.
\newblock Reinforcement learning-based black-box model inversion attacks.
\newblock In \emph{Proceedings of the IEEE/CVF Conference on Computer Vision and Pattern Recognition}, pages 20504--20513.

\bibitem[{Hendrycks et~al.(2020)Hendrycks, Burns, Basart, Zou, Mazeika, Song, and Steinhardt}]{hendrycks2020measuring}
Dan Hendrycks, Collin Burns, Steven Basart, Andy Zou, Mantas Mazeika, Dawn Song, and Jacob Steinhardt. 2020.
\newblock Measuring massive multitask language understanding.
\newblock \emph{arXiv preprint arXiv:2009.03300}.

\bibitem[{Honovich et~al.(2022)Honovich, Shaham, Bowman, and Levy}]{honovich2022instruction}
Or~Honovich, Uri Shaham, Samuel~R Bowman, and Omer Levy. 2022.
\newblock Instruction induction: From few examples to natural language task descriptions.
\newblock \emph{arXiv preprint arXiv:2205.10782}.

\bibitem[{Hou et~al.(2023)Hou, O’connor, Andreas, Chang, and Zhang}]{hou2023promptboosting}
Bairu Hou, Joe O’connor, Jacob Andreas, Shiyu Chang, and Yang Zhang. 2023.
\newblock Promptboosting: Black-box text classification with ten forward passes.
\newblock In \emph{International Conference on Machine Learning}, pages 13309--13324. PMLR.

\bibitem[{Hu et~al.(2021)Hu, Shen, Wallis, Allen-Zhu, Li, Wang, Wang, and Chen}]{hu2021lora}
Edward~J Hu, Yelong Shen, Phillip Wallis, Zeyuan Allen-Zhu, Yuanzhi Li, Shean Wang, Lu~Wang, and Weizhu Chen. 2021.
\newblock Lora: Low-rank adaptation of large language models.
\newblock \emph{arXiv preprint arXiv:2106.09685}.

\bibitem[{Jiang et~al.(2023)Jiang, Sablayrolles, Mensch, Bamford, Chaplot, Casas, Bressand, Lengyel, Lample, Saulnier et~al.}]{jiang2023mistral}
Albert~Q Jiang, Alexandre Sablayrolles, Arthur Mensch, Chris Bamford, Devendra~Singh Chaplot, Diego de~las Casas, Florian Bressand, Gianna Lengyel, Guillaume Lample, Lucile Saulnier, et~al. 2023.
\newblock Mistral 7b.
\newblock \emph{arXiv preprint arXiv:2310.06825}.

\bibitem[{Lester et~al.(2021)Lester, Al-Rfou, and Constant}]{lester2021power}
Brian Lester, Rami Al-Rfou, and Noah Constant. 2021.
\newblock The power of scale for parameter-efficient prompt tuning.
\newblock \emph{arXiv preprint arXiv:2104.08691}.

\bibitem[{Liu et~al.(2019)Liu, Ott, Goyal, Du, Joshi, Chen, Levy, Lewis, Zettlemoyer, and Stoyanov}]{liu2019roberta}
Yinhan Liu, Myle Ott, Naman Goyal, Jingfei Du, Mandar Joshi, Danqi Chen, Omer Levy, Mike Lewis, Luke Zettlemoyer, and Veselin Stoyanov. 2019.
\newblock Roberta: A robustly optimized bert pretraining approach.
\newblock \emph{arXiv preprint arXiv:1907.11692}.

\bibitem[{Mihaylov et~al.(2018)Mihaylov, Clark, Khot, and Sabharwal}]{OpenBookQA2018}
Todor Mihaylov, Peter Clark, Tushar Khot, and Ashish Sabharwal. 2018.
\newblock Can a suit of armor conduct electricity? a new dataset for open book question answering.
\newblock In \emph{EMNLP}.

\bibitem[{Mishra et~al.(2022)Mishra, Khashabi, Baral, and Hajishirzi}]{naturalinstructions}
Swaroop Mishra, Daniel Khashabi, Chitta Baral, and Hannaneh Hajishirzi. 2022.
\newblock Cross-task generalization via natural language crowdsourcing instructions.
\newblock In \emph{ACL}.

\bibitem[{Ouyang et~al.(2022)Ouyang, Wu, Jiang, Almeida, Wainwright, Mishkin, Zhang, Agarwal, Slama, Ray et~al.}]{ouyang2022training}
Long Ouyang, Jeffrey Wu, Xu~Jiang, Diogo Almeida, Carroll Wainwright, Pamela Mishkin, Chong Zhang, Sandhini Agarwal, Katarina Slama, Alex Ray, et~al. 2022.
\newblock Training language models to follow instructions with human feedback.
\newblock \emph{Advances in neural information processing systems}, 35:27730--27744.

\bibitem[{Prasad et~al.(2022)Prasad, Hase, Zhou, and Bansal}]{prasad2022grips}
Archiki Prasad, Peter Hase, Xiang Zhou, and Mohit Bansal. 2022.
\newblock Grips: Gradient-free, edit-based instruction search for prompting large language models.
\newblock \emph{arXiv preprint arXiv:2203.07281}.

\bibitem[{Pryzant et~al.(2023)Pryzant, Iter, Li, Lee, Zhu, and Zeng}]{pryzant2023automatic}
Reid Pryzant, Dan Iter, Jerry Li, Yin~Tat Lee, Chenguang Zhu, and Michael Zeng. 2023.
\newblock Automatic prompt optimization with" gradient descent" and beam search.
\newblock \emph{arXiv preprint arXiv:2305.03495}.

\bibitem[{Radford et~al.(2019)Radford, Wu, Child, Luan, Amodei, Sutskever et~al.}]{radford2019language}
Alec Radford, Jeffrey Wu, Rewon Child, David Luan, Dario Amodei, Ilya Sutskever, et~al. 2019.
\newblock Language models are unsupervised multitask learners.
\newblock \emph{OpenAI blog}, 1(8):9.

\bibitem[{Schulman et~al.(2017)Schulman, Wolski, Dhariwal, Radford, and Klimov}]{schulman2017proximal}
John Schulman, Filip Wolski, Prafulla Dhariwal, Alec Radford, and Oleg Klimov. 2017.
\newblock Proximal policy optimization algorithms.
\newblock \emph{arXiv preprint arXiv:1707.06347}.

\bibitem[{Team et~al.(2024)Team, Mesnard, Hardin, Dadashi, Bhupatiraju, Pathak, Sifre, Rivi{\`e}re, Kale, Love et~al.}]{team2024gemma}
Gemma Team, Thomas Mesnard, Cassidy Hardin, Robert Dadashi, Surya Bhupatiraju, Shreya Pathak, Laurent Sifre, Morgane Rivi{\`e}re, Mihir~Sanjay Kale, Juliette Love, et~al. 2024.
\newblock Gemma: Open models based on gemini research and technology.
\newblock \emph{arXiv preprint arXiv:2403.08295}.

\bibitem[{Touvron et~al.(2023)Touvron, Lavril, Izacard, Martinet, Lachaux, Lacroix, Rozi{\`e}re, Goyal, Hambro, Azhar et~al.}]{touvron2023llama}
Hugo Touvron, Thibaut Lavril, Gautier Izacard, Xavier Martinet, Marie-Anne Lachaux, Timoth{\'e}e Lacroix, Baptiste Rozi{\`e}re, Naman Goyal, Eric Hambro, Faisal Azhar, et~al. 2023.
\newblock Llama: Open and efficient foundation language models.
\newblock \emph{arXiv preprint arXiv:2302.13971}.

\bibitem[{Wang et~al.(2019)Wang, Pruksachatkun, Nangia, Singh, Michael, Hill, Levy, and Bowman}]{wang2019superglue}
Alex Wang, Yada Pruksachatkun, Nikita Nangia, Amanpreet Singh, Julian Michael, Felix Hill, Omer Levy, and Samuel Bowman. 2019.
\newblock Superglue: A stickier benchmark for general-purpose language understanding systems.
\newblock \emph{Advances in neural information processing systems}, 32.

\bibitem[{Wang et~al.(2018)Wang, Singh, Michael, Hill, Levy, and Bowman}]{wang2018glue}
Alex Wang, Amanpreet Singh, Julian Michael, Felix Hill, Omer Levy, and Samuel~R Bowman. 2018.
\newblock Glue: A multi-task benchmark and analysis platform for natural language understanding.
\newblock \emph{arXiv preprint arXiv:1804.07461}.

\bibitem[{Wang et~al.(2023)Wang, Li, Wang, Bai, Luo, Zhang, Jojic, Xing, and Hu}]{wang2023promptagent}
Xinyuan Wang, Chenxi Li, Zhen Wang, Fan Bai, Haotian Luo, Jiayou Zhang, Nebojsa Jojic, Eric~P Xing, and Zhiting Hu. 2023.
\newblock Promptagent: Strategic planning with language models enables expert-level prompt optimization.
\newblock \emph{arXiv preprint arXiv:2310.16427}.

\bibitem[{Wei et~al.(2022)Wei, Wang, Schuurmans, Bosma, Xia, Chi, Le, Zhou et~al.}]{wei2022chain}
Jason Wei, Xuezhi Wang, Dale Schuurmans, Maarten Bosma, Fei Xia, Ed~Chi, Quoc~V Le, Denny Zhou, et~al. 2022.
\newblock Chain-of-thought prompting elicits reasoning in large language models.
\newblock \emph{Advances in neural information processing systems}, 35:24824--24837.

\bibitem[{Yao et~al.(2022)Yao, Chen, Yang, and Narasimhan}]{yao2022webshop}
Shunyu Yao, Howard Chen, John Yang, and Karthik Narasimhan. 2022.
\newblock Webshop: Towards scalable real-world web interaction with grounded language agents.
\newblock \emph{Advances in Neural Information Processing Systems}, 35:20744--20757.

\bibitem[{Zeng et~al.(2022)Zeng, Liu, Du, Wang, Lai, Ding, Yang, Xu, Zheng, Xia et~al.}]{zeng2022glm}
Aohan Zeng, Xiao Liu, Zhengxiao Du, Zihan Wang, Hanyu Lai, Ming Ding, Zhuoyi Yang, Yifan Xu, Wendi Zheng, Xiao Xia, et~al. 2022.
\newblock Glm-130b: An open bilingual pre-trained model.
\newblock \emph{arXiv preprint arXiv:2210.02414}.

\bibitem[{Zhang et~al.(2022{\natexlab{a}})Zhang, Roller, Goyal, Artetxe, Chen, Chen, Dewan, Diab, Li, Lin et~al.}]{zhang2022opt}
Susan Zhang, Stephen Roller, Naman Goyal, Mikel Artetxe, Moya Chen, Shuohui Chen, Christopher Dewan, Mona Diab, Xian Li, Xi~Victoria Lin, et~al. 2022{\natexlab{a}}.
\newblock Opt: Open pre-trained transformer language models.
\newblock \emph{arXiv preprint arXiv:2205.01068}.

\bibitem[{Zhang et~al.(2022{\natexlab{b}})Zhang, Wang, Zhou, Schuurmans, and Gonzalez}]{zhang2022tempera}
Tianjun Zhang, Xuezhi Wang, Denny Zhou, Dale Schuurmans, and Joseph~E Gonzalez. 2022{\natexlab{b}}.
\newblock Tempera: Test-time prompting via reinforcement learning.
\newblock \emph{arXiv preprint arXiv:2211.11890}.

\bibitem[{Zhou et~al.(2022)Zhou, Muresanu, Han, Paster, Pitis, Chan, and Ba}]{zhou2022large}
Yongchao Zhou, Andrei~Ioan Muresanu, Ziwen Han, Keiran Paster, Silviu Pitis, Harris Chan, and Jimmy Ba. 2022.
\newblock Large language models are human-level prompt engineers.
\newblock \emph{arXiv preprint arXiv:2211.01910}.

\end{thebibliography}
\appendix

\section{Experiment Details}
  
\subsection{RL parameters}
We summarize the RL-related hyperparameters used in our experiments in the ~\Cref{tab:RL_parameter}. We used the same hyperparameters for all tasks.

\subsection{Meta Prompt}
\begin{figure}[t]
    \centerline{\includegraphics[width=\columnwidth]{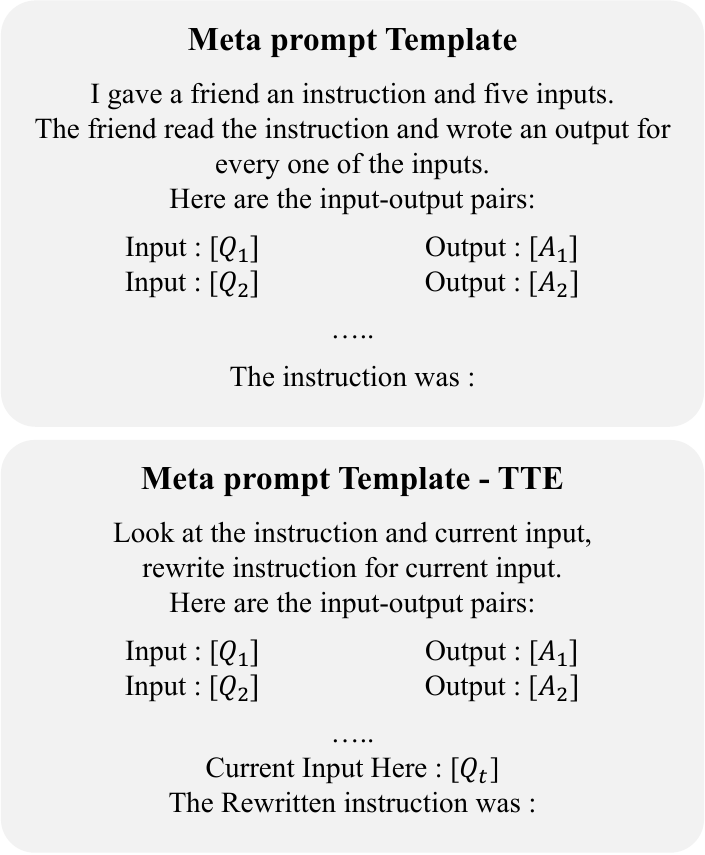}}
    \caption{Detail template of meta prompt used in StablePrompt and TTE-StablePrompt}
    \label{Fig:meta-prompt} 
\end{figure}

We show the meta-prompt used as input to StablePrompt in ~\Cref{Fig:meta-prompt}.

\begin{table}[]
\centering
\resizebox{1.0\columnwidth}{!}{%
\begin{tabular}{@{}ll@{}}
\toprule
HyperParameters        & \multicolumn{1}{l}{Stableprompt} \\ \midrule
Learning Rate          & 1.00E-05                         \\
Value loss Coefficient & 0.1                              \\
Gamma                  & 1                                \\
GAE Lambda             & 0.95                             \\
cliprange              & 0.2                              \\ 
$u_t$            & 5                                \\
Update Threshold(\%)       & 0.05                            \\
Rollback Threshold(\%) &0.1\\
Prompt per Batch       & 4                                  \\
Maximum Prompt Length  & 150                                \\
$c_a$                     & 10                                \\
$c_s$                     & 0.1                               \\
\bottomrule
\end{tabular}%
}
\caption{Detail parameters used in StablePrompt.}
\label{tab:RL_parameter}
\end{table}

\subsection{Dataset Details}
\begin{table}[h]
\resizebox{\columnwidth}{!}{%
\begin{tabular}{@{}ccccccc@{}}
\toprule
Dataset & Type & |C| & |Train|=|Dev| & |Test| & Verbalizer  \\ \midrule
SST2 & sentiment & 2 & 32 & 1.8k & [yes,no]      \\
MRPC  & NLI       & 2 & 32 & 1.7k   & [yes,no]       \\
RTE   & NLI       & 2 & 32 & 0.3k & [yes,no]              \\
QNLI  & NLI       & 2 & 32 & 9.8k & [yes,no]             \\
MNLI  & NLI       & 3 & 48 & 10k  & [yes,maybe,no]  \\
SNLI  & NLI       & 3 & 48 & 9.8k & [yes,maybe,no]   \\ 
MMLU  & QA        & 4 & -  & -  & [A,B,C,D] \\

\bottomrule
\end{tabular}%
}
\caption{Details of the datasets for few-shot classification.}
\label{tab:appendix_fewshot-dataset-detail}
\end{table}
\paragraph{Few-shot Text Classification} \label{appendix_detail_Fewshot}
Detailed number and verbalizer settings can be found in~\Cref{tab:appendix_fewshot-dataset-detail}.

\paragraph{Induction Task} \label{appendix_detail_Induction}
BIG-Bench Instruction Induction (BBII) is a subset of 21 tasks with clear and human-written instructions that can be applied to all examples in the dataset~\cite{zhou2022large}. The detailed type and metric for each dataset can be found in~\Cref{tab:appendix_BBH_full}.

Instruction Induction is conducted with 24 induction tasks proposed in ~\cite{honovich2022instruction}. The tasks span many features of language understanding, from
simple phrase structure to similarity and causality identification.
The detailed metric for each dataset can be found in~\Cref{tab:appendix_ii_full}.

\paragraph{Question Answering} \label{appendix_detail_qa}
The MMLU QA dataset consists of 15,908 questions. The dataset is divided into subsets according to 57 subjects. We use the validation set of all subsets as the training set. The total number of validation sets is 1,540. Each subset has a minimum of 100 test samples, with a total of 14,079 test questions.

\subsection{Baseline Details}
\paragraph{APE}
For a fair comparison, we scale the number of prompts generated by APE to be the same as the number that StablePrompt generates during training. Also, unlike the original APE, we use the entire validation set to determine the final prompt. This setting is more favorable than the original APE and improves performance.
\paragraph{ProTeGi}
We use additional settings same as APE and limited the number of consecutive conversations to two.
\paragraph{RLprompt}
For RLprompt, as the hidden size of the agent model increases, the size of the MLP layer increases as well, making it difficult to train the model. 
Therefore, we use GPT2-XL~\cite{radford2019language} 1.5B, which is the largest model in the official implementation.

\paragraph{PromptAgent}
We utilize the official repository and only used it for the text classification problem as no evaluation metric was specified for text generation.
PromptAgent is known to work well on high-performance LLMs such as GPT-4. However, in our experiments, we found that using small 7B-level models as agents significantly degrades performance.

\subsection{Training Details}
We experiment on a single A100 GPU. For text classification, we use 100 epochs and need 2-3 GPU hours per task. For question-answering and induction tasks, we use 30 epochs and need 1-2 GPU hours per task. Training time can be changed by the average length of inputs.

\section{Additional Experiments}

\subsection{Text Classification in Small Target Model}
\paragraph{Implementation Details}
To compare the performance of our methods with traditional prompt tuning baselines, we perform text classification again on a relatively smaller target model. The target model is fixed as RoBERTa-Large (330M). 
We replace the MRPC dataset with the MR dataset from glue. Note that the MR dataset is a sentiment classification task, not NLI.
For RLprompt, the agent model is GPT-2, as specified in the original paper. For StablePrompt, the agent model used is Mistral-7B.

\begin{table*}[!h]
    \resizebox{\textwidth}{!}{%
    \begin{tabular}{@{}cccccccc@{}}
    \toprule
                                     &                 & SST2           & MR             & RTE            & QNLI           & MNLI           & SNLI           \\ \midrule
    Fine-Tuning                      & Fine-Tuning      & 80.4($\pm$3.9) & 67.4($\pm$9.7) & 58.6($\pm$3.9) & 60.2($\pm$4.7) & 47.8($\pm$7.5) & 54.6($\pm$9.7) \\ \midrule
    \multirow{2}{*}{Continous prompt} &
      Soft prompt Tuning &
      73.8($\pm$10.9) &
      88.6($\pm$14.6) &
      54.7($\pm$10.9) &
      49.7($\pm$0.2) &
      33.2($\pm$0.0) &
      36.1($\pm$14.6) \\
                                     & Blackbox-Tuning & 89.1($\pm$0.9) & 93.2($\pm$1.3) & 52.6($\pm$0.9) & 48.8($\pm$0.6) & 42.9($\pm$2.0) & 46.6($\pm$1.3) \\ \midrule
    \multirow{6}{*}{Discrete prompt} & Manual Prompt   & 82.8           & 80.9           & 51.6           & 50.8           & 51.7           & 31.1           \\
                                     
                                     & In-Context Demo & 85.9($\pm$0.7) & 80.6($\pm$1.4) & 60.4($\pm$0.7) & 53.8($\pm$0.4) & 53.4($\pm$1.5) & 47.1($\pm$1.4) \\
                                     & GrIPS           & 87.1($\pm$1.5) & 80.0($\pm$2.5) & 48.6($\pm$1.0) & 50.4($\pm$0.4) & 35.2($\pm$0.3) & 33.3($\pm$0.0) \\
                                     & PromptBoosting  & 89.8($\pm$1.1) & 86.0($\pm$3.5) & 57.2($\pm$2.7) & 56.9($\pm$2.1) & 43.8($\pm$1.1) & 53.6($\pm$3.3) \\
                                     & APE             & 82.5($\pm$4.7) & 82.8($\pm$4.7) & 57.3($\pm$4.1) & 54.5($\pm$3.2) & 45.6($\pm$1.8) & 49.6($\pm$3.5) \\
                                     & RLprompt        & 90.1($\pm$1.8) & 86.7($\pm$2.4) & 50.2($\pm$3.1) & 33.3($\pm$0.0) & 35.0($\pm$0.4) & 32.1($\pm$0.2) \\ \midrule
    Test-time editing                & tempera         & 91.9($\pm$2.0) & \textbf{88.0}($\pm$1.1) & 60.3($\pm$2.2) & 57.4($\pm$1.5) & 45.2($\pm$2.0) & \textbf{56.4}($\pm$3.2) \\ \midrule
    Discrete prompt &
      \begin{tabular}[c]{@{}c@{}}Stableprompt\\ (Ours)\end{tabular} &
      \textbf{92.8}($\pm$0.8) &
      87.4($\pm$0.1) &
      \textbf{62.9}($\pm$0.8) &
      \textbf{59.1}($\pm$0.6) &
      \textbf{49.1}($\pm$2.6) &
      55.3($\pm$0.9) \\ \bottomrule
    \end{tabular}%
    }
    \caption{Mean and standard deviation of accuracy on three random seeds of the few-shot text classification task on the roberta-large (330M) target model setting.}
    \label{tab:roberta-fewshot}
\end{table*}

\paragraph{Results}
The experimental results are shown in ~\Cref{tab:roberta-fewshot}. StablePrompt demonstrates the highest performance across all datasets except MR. Even in MR, it shows comparable performance to TEMPERA, which uses Test-Time Editing for multi-step reasoning, thereby proving the high performance of our model.

\subsection{Ablation Study} \label{appendix_ablation}
\begin{figure}[t]
    \centerline{\includegraphics[width=\columnwidth]{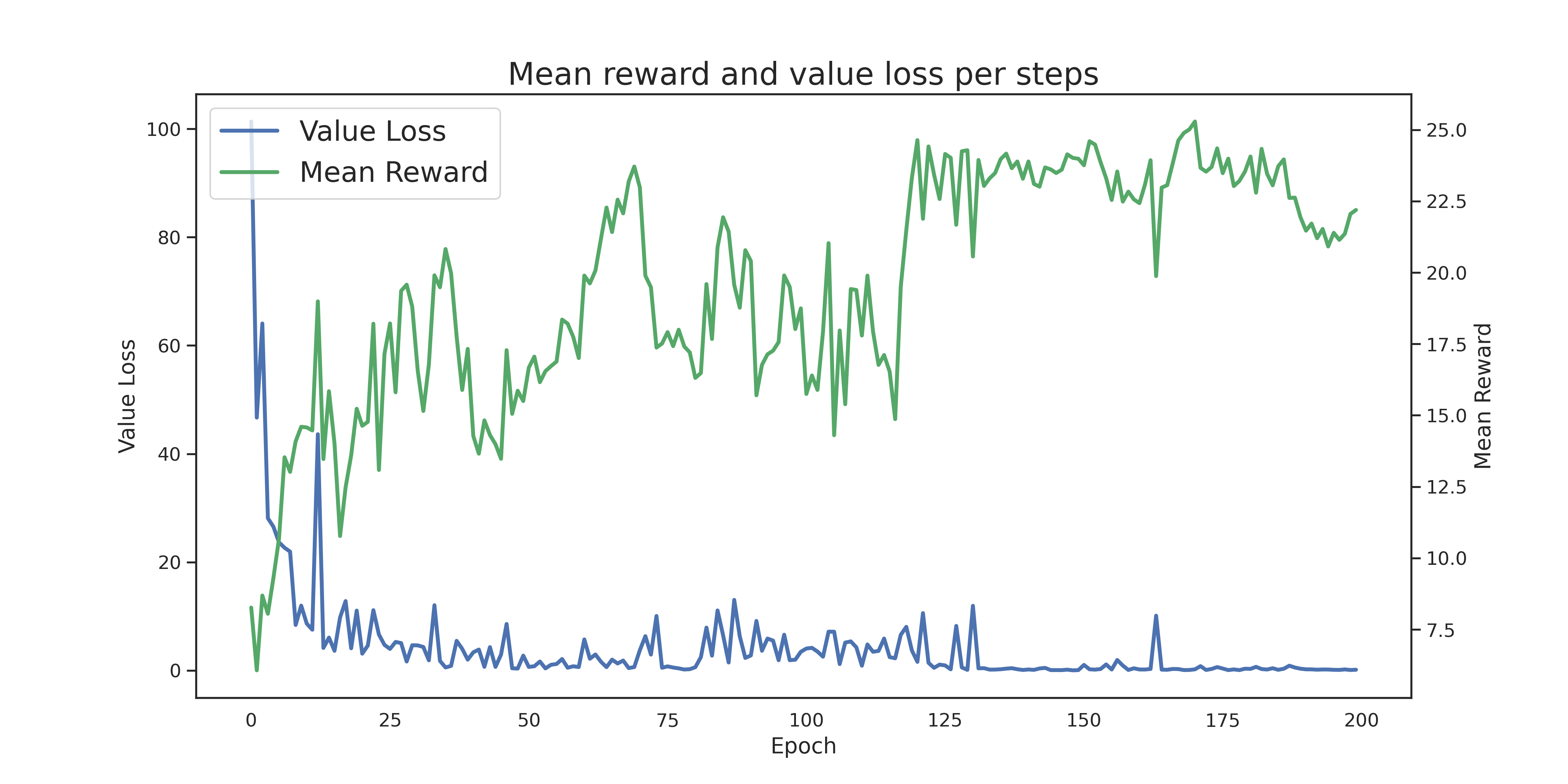}}
    \caption{Training curve of mean reward and value loss by steps.}
    \label{Fig:Reward} 
\end{figure}
\paragraph{Training curve}
~\Cref{Fig:Reward} shows the mean reward and value loss by steps. We experiment on the few-shot text classification task with the SST2 dataset. This shows a steady increase in reward, indicating that our method is training well. In addition, value loss, the MSE loss of the reward expected by the value head, also falls steadily over time. This shows that the value head is also aligned with the reward model.

\begin{table}[]
\centering
\resizebox{0.8\columnwidth}{!}{%
\begin{tabular}{@{}ccc@{}}
\toprule
Dataset                   & SST2          \\ \midrule             
Ours               & \textbf{94.6}($\pm$0.6)   \\
w/o softmax difference  & 93.31($\pm$0.8) \\ \bottomrule
\end{tabular}%
}
\caption{Ablation study of reward function terms on sst2 dataset.}
\label{tab:reward-ablation}
\end{table}
\paragraph{Reward Function Ablation}
~\Cref{tab:reward-ablation} shows an ablation study for reward function. We use the same setting as ~\cref{few-shot}, but change the agent and target model to Mistral-7B. For text generation, we use only the f1 score. The continuous value of the f1 score is proper for the reward function. But in text classification, we introduce softmax difference. A smaller batch size will result in many prompts with the same accuracy, which will confuse the model. To avoid this, softmax difference ranks prompts with the same accuracy. In practice, not using softmax difference results in a performance penalty.

\section{Generated Prompt} 
We attach the generated prompts below, which we can not show on the page. For each task, we post one of the best-performing prompts.
\subsection{Few-Shot Text Classification} \label{Generated_Prompt_FSTC}
\paragraph{SST2}
**Write yes or no for each input, based on whether the input describes a movie that would be reasonably entertaining or not.** **Input 1:** Reasonably entertaining sequel **Output:** yes **Input 2:** Familiar and predictable, and 4/5ths of it **Output:** no

\paragraph{MRPC}
**Write "Yes" or "No" for each sentence pair, based on whether the second sentence is a paraphrase of the first sentence.**
**Here are the outputs:**
- Sentence1 : The woman was hospitalized June 15 , Kansas health officials said .
Sentence2 : Missouri health officials said he had not been hospitalized and is recovering .
**Output:** No
- Sentence1 : CS 's other main division , Financial Services , made a 666 million franc net profit , six percent below the prior quarter .
Sentence2 : CS Financial Services made a 666 million franc net profit , six percent less than in the fourth quarter of last year .
**Output:** Yes
- Sentence1 : It has been named Colymbosathon ecplecticos , which means " astounding swimmer with a large penis " .
Sentence2 : He and colleagues named it Colymbosathon ecplecticos , which means " swimmer with a

\paragraph{RTE}
**For each input-output pair:**
* Carefully read the premise.
* Consider the relationship between the premise and the hypothesis.
* Based on the information provided, determine whether the output (yes/no) is consistent with the premise and hypothesis.
* Provide your reasoning and explanation for your answer.

\paragraph{QNLI}
**Given a question and a sentence, determine whether the sentence provides evidence that the statement in the question is true or false.**
From the input-output pairs, it can be observed that your friend has a correct understanding of the instruction. They correctly identified whether the given sentence provides evidence to support the claim in the question for each of the five inputs.
\paragraph{MNLI}
**Step 1: Analyze the premise**
- Carefully read the given premise.
- Identify the main points mentioned.
- Understand the emotional tone or sentiment expressed.
**Step 2: Consider the hypothesis**
- Examine the proposed hypothesis.
- Determine the reasoning behind it.
- Identify the evidence or logic supporting it.
**Step 3: Predic the output**
- Based on your understanding of the premise and hypothesis, predict the likely output.
\paragraph{SNLI}
Imagine you're given some information about a scene, like a sentence describing what's happening.
Your job is to analyze the information and predict whether the provided hypothesis is true or false based on the given premise.
For example:
**Premise:** A dog catches a disk in the air.
**Hypothesis:** A dog is eating kibble out of a red bowl.
**Output:** No
Remember to carefully analyze the details of the scene and consider how they relate to the hypothesis

\subsection{BigBench-Hard Insstruction Induction} \label{Generated_Prompt_BBH}
We choose random 3 tasks from the BBH-II dataset.
\paragraph{Causal Judgment}
For each input, write an output indicating whether the person intentionally obtained the item in question.
From the outputs, we can see that:
**Input 1:** Joe intentionally did not want the commemorative cup, despite being offered it.
**Input 2:** Professor Smith intentionally took the pen despite knowing it was only for administrative assistants.
Therefore, both outputs indicate that the people intentionally obtained the items they received.
\paragraph{Navigate}
**Write True or False based on the following statement:**
"If you always face forward and take an even number of steps, you will end up at the same starting point."
**The outputs shows that the statement is True for all of the input except for the last one.**
Therefore, the answer is **B : False.**
\paragraph{Tense}
Change each sentence given into the Past Perfect Tense.
**The provided input-output pairs are all correct.**
**Here are the sentences in Past Perfect Tense:**
1. She had washed her hands before she cooked the dinner.
2. The weather changed, but the team had planned its next move.
3. The girl bought the telescope her teacher had recommended to her.
4. I had finished the meal.
5. Raj had read a novel.

\subsection{Instruction Induction} \label{Generated_Prompt_II}
We choose random 3 tasks from the II dataset.
\paragraph{Negation}
For each of the following statements, simply negate the main clause.
**Here are the corrected outputs:**
**1.** Fritschel said the man's sleep was disturbed by dreams.
**Output:** Fritschel did not say the man's sleep was disturbed by dreams.
**2.** In terms of most populous counties in the United States the lowest ranking is Riverside.
**Output:** In terms of most populous counties in the United States the lowest ranking is not Riverside.
\paragraph{Num to Verbal}
**Write the word form of the given number for each input.**
The outputs are correct for all the inputs.
**Input :** 87
**Output:** Eighty-seven
**Input :** 75
**Output:** Seventy-five
**Input :** 77
**Output:** Seventy-seven
**Input :** 88
**Output:** Eighty-eight
**Input :** 82
**Output:** Eighty-two
\paragraph{Taxonomy Animal}
**For each input, write the three most common animals from the input.**
**Here are the outputs:**
**1. sock, helicopter, dolphin, cat, shirt, lion**
 - Dolphin, cat, lion
**2. motorway, truck, octopus, dress, fly, tortoise**
 - Fly, octopus, tortoise
**3. lizard, dolphin, nurse, goat, white, gold**
 - Goat, lizard, dolphin
**4. kiwi, bat, gray

\subsection{Question Answering} \label{Generated_Prompt_QA}
We choose random 4 subjects from MMLU.
\paragraph{Moral Scenarios}
Look at these input-output pairs. Each pair consists of a scenario and a judgment about whether the main character in the scenario does something clearly morally wrong, according to ordinary moral standards in the US as of 2020. Try to understand the logic behind each judgment. Can you guess the output for the next scenario, based on the patterns you've identified?

\paragraph{Machine Learning}
**For each input-output pair:**
1. **Look at the input carefully.** Identify the key terms, concepts, or characteristics.
2. **Consider the output.** What does it suggest about the relationship between the input and the desired result?
3. **Reflect on your knowledge** of related concepts and theories. Identify any connections between the input and output.
4. **Make an educated guess** about the output based on your understanding of the input and the relationships you identified.
\paragraph{High School Biology}
 **For each question:**
- Carefully read the input information.
- Identify the main idea or central detail in the input.
- Consider the available options and think about which one best matches the input.
- Select the option that aligns with the information given in the input.
Remember to take your time and focus on understanding the input before making your guess.
\paragraph{College Chemistry}
**For each input-output pair:**
1. **Analyze the input:** Carefully review the given information about the elements or the chemical compound. Look for any relevant properties or characteristics.
2. **Consider the possibilities:** Recall the different options given in the choices (A, B, C, D).
3. **Apply your knowledge:** Use your understanding of the relevant concepts or principles to determine which option is most likely to be the correct output.
4. **Check your reasoning:**

\subsection{Full prompt of Figure 5} \label{appendix_figure5}
\paragraph{Question}
question : In building a linear regression model for a particular data set, you observe the coefficient of one of the features having a relatively high negative value. This suggests that

\paragraph{Choices}
choices :
 A : This feature has a strong effect on the model (should be retained)
B : This feature does not have a strong effect on the model (should be ignored)
C : It is not possible to comment on the importance of this feature without additional information
D : Nothing can be determined.
\paragraph{APE}
For each input-output pair provided, carefully read the statement and consider the available choices. Based on your understanding of the information presented, choose the option that best aligns with the given output. Remember to analyze the statements, identify key concepts, and apply your knowledge to make an informed guess about the correct answer 
\paragraph{StablePrompt}
**Instructions:**
 Look at the input and try to understand what information is being presented.
 Consider the possible outputs based on the input characteristics.
Choose the most likely

\paragraph{TTE-StablePrompt}
Remember the bias-variance trade-off in model building. Consider whether a high negative coefficient for a feature might indicate that the model is overly tuned to the training set (high variance) or truly capturing meaningful information from the feature.

\section{Full Experiment Results}
\subsection{Question Answering}
We show the performance on the entire QA dataset in the~\Cref{tab:appendix_mmlu_full}.

\subsection{Instruction Induction}
We show the performance on the BBH-II and II full datasets in the~\Cref{tab:appendix_BBH_full} and~\Cref{tab:appendix_ii_full}.
We also show the results for APE, StablePrompt, and human prompt for both the Target and Agent models using InstructGPT3.5.

\begin{table*}[]
\resizebox{\textwidth}{!}{%
\begin{tabular}{@{}lllccccccc@{}}
\toprule
task name &
  type &
  Metric &
  fewshot &
  manual &
  APE &
  ProTeGi &
  PromptAgent &
  StablePrompt \\ \midrule
causal judgment &
  Multiple Choice &
  Accuracy &
  \textbf{58.75} &
  52.50 &
  {\ul 58.13} &
  56.69 &
  31.65 &
  \textbf{58.75} \\
disambiguation qa &
  Multiple Choice &
  Accuracy &
  \textbf{64.29} &
  52.19 &
  64.00 &
  61.40 &
  2.54 &

  {\ul 64.04} \\
epistemic reasoning &
  Multiple Choice &
  Accuracy &
  43.69 &
  57.16 &
  58.40 &
  \textbf{63.79} &
  13.92 &

  {\ul 61.47} \\
hyperbaton &
  Multiple Choice &
  Accuracy &
  47.89 &
  56.52 &
  {\ul 75.60} &
  \textbf{76.06} &
  56.96 &

  {\ul 75.60} \\
implicatures &
  Multiple Choice &
  Accuracy &
  \textbf{83.33} &
  {\ul 83.12} &
  80.95 &
  73.59 &
  55.70 &

  79.00 \\
logical fallacy detection &
  Multiple Choice &
  Accuracy &
  58.19 &
  \textbf{63.50} &
  56.50 &
  58.23 &
  37.97 &

  {\ul 58.34} \\
movie recommendation &
  Multiple Choice &
  Accuracy &
  49.36 &
  37.66 &
  {\ul 55.30} &
  \textbf{67.23} &
  22.78 &

  {\ul 55.30} \\
navigate &
  Multiple Choice &
  Accuracy &
  \textbf{69.22} &
  49.79 &
  52.28 &
  {\ul 54.02} &
  35.44 &

  53.30 \\
presuppositions as nli &
  Multiple Choice &
  Accuracy &
  42.55 &
  40.82 &
  41.56 &
  41.42 &
  0.00 &

  \textbf{43.40} \\
ruin names &
  Multiple Choice &
  Accuracy &
  12.44 &
  30.14 &
  {\ul 32.53} &
  27.99 &
  21.52 &

  \textbf{37.08} \\
snarks &
  Multiple Choice &
  Accuracy &
  35.79 &
  42.38 &
  {\ul 50.99} &
  {\ul 50.99} &
  0.00 &

  \textbf{52.32} \\
sportsunderstanding &
  Multiple Choice &
  Accuracy &
  52.37 &
  {\ul 59.38} &
  56.50 &
  55.98 &
  2.00 &

  \textbf{60.12} \\
dyck languages &
  Generation &
  Exact Match &
  0.00 &
  0.00 &
  0.00 &
  0.00 &
  - &

  0.00 \\
\begin{tabular}[c]{@{}l@{}}gender inclusive\\ sentences german\end{tabular} &
  Generation &
  Exact Match &
  9.30 &
  86.00 &
  67.13 &
  \textbf{93.77} &
  - &

  89.70 \\
object counting &
  Generation &
  Exact Match &
  7.13 &
  0.00 &
  14.29 &
  \textbf{33.33} &
  - &

  {\ul 15.71} \\
operators &
  Generation &
  Exact Match &
  5.53 &
  49.45 &
  {\ul 57.14} &
  50.00 &
  - &

  \textbf{64.29} \\
tense &
  Generation &
  Exact Match &
  15.29 &
  93.85 &
  96.76 &
  \textbf{100.00} &
  - &

  {\ul 98.43} \\
word sorting &
  Generation &
  Exact Match &
  0.00 &
  20.14 &
  {\ul 96.43} &
  75.00 &
  - &

  \textbf{100.00} \\ \bottomrule
\end{tabular}%
}
\caption{Results of full experiment of BigBench-Hard Instruction Induction datasets with Gemma-7B as target model.}
\label{tab:appendix_BBH_full}
\end{table*}
\begin{table*}[]
\resizebox{\textwidth}{!}{%
\begin{tabular}{@{}llcccccc@{}}
\toprule
taskname                 & Metric     & fewshot & manual  & APE        & ProTeGi         & StablePrompt    \\ \midrule
antonyms                & Exact Match & 0       & 0.43    & 0.625      & 0.25            & \textbf{0.75}   \\
word in context         & Exact Match & 0.55    & 0.46    & 0.375      & 0.5                       & \textbf{0.8125} \\
rhymes                  & Exact Match & 0       & 0.03    & 0.0625     & 0.25           & 0.0625          \\
num to verbal           & Exact Match & 0       & 0.61    & 0.9375     & \textbf{1}         & \textbf{1}      \\
cause and effect        & Exact Match & 0       & 0.24    & 0.6        & 0                        & \textbf{0.7}    \\
larger animal           & Exact Match & 0       & 0.03    & 0.5625     & 0.25                  & \textbf{0.9375} \\
second word letter      & Exact Match & 0.12    & 0.08    & 0.0625     & 0.25         & \textbf{0.1875} \\
taxonomy animal         & Exact Set   & 0       & 0       & 0.375      & 0.375                 & \textbf{0.5}    \\
negation                & Exact Match & 0       & 0.16    & 0.6875     & 0.5                   & \textbf{0.75}   \\
common concept          & F1          & 0.03    & 0.04    & 0.5        & 0.5             & 0.75            \\
diff                    & Exact Match & 0.02    & 0.99    & \textbf{1} & \textbf{1}        & \textbf{1}      \\
translation en-es       & Exact Match & 0       & 0.15    & 0.25       & 0.25                 & 0.4375          \\
orthography starts with & Exact Set   & 0       & 0.375   & 0.125      & 0                    & 0.375           \\
sentiment               & Exact Match & 0.5     & 0.83    & 0.6875     & 1               & 1               \\
informal to formal      & F1          & 0       & 0.27384 & 0.425      & 0.2422            & 0.4641          \\
sum                     & Exact Match & 0       & 0.99    & 1          & 1                 & 1               \\
singular to plural      & Exact Match & 0       & 0.75    & 0.9375     & 1                  & 1               \\
active to passive       & Exact Match & 0       & 0.53    & 1          & 1                  & 1               \\
translation en-de       & Exact Match & 0       & 0.1     & 0.1875     & \textbf{0.5}         & 0.3125          \\
sentence similarity     & Exact Match & 0       & 0.2     & 0.315      & 0.25            & \textbf{0.5}    \\
translation en-fr       & Exact Match & 0       & 0.07    & 0.06       & 0.5          & 0.315           \\
letters list            & Exact Match & 0       & 0       & 0.6875     & 0.5           & \textbf{0.875}  \\
first word letter       & Exact Match & 0.03    & 0.73    & 0.8775     & 1                   & \textbf{0.9375} \\
synonyms                & Contains    & 0       & 0.02    & 0.125      & 0.25          & 0.125           \\ \bottomrule
\end{tabular}%
}
\caption{Results of full experiment of Instruction Induction datasets with Gemma-7B as target model.}
\label{tab:appendix_ii_full}
\end{table*}
\begin{table*}[]
\centering
\resizebox{0.7\textwidth}{!}{%
\begin{tabular}{@{}ccccc@{}}
\toprule
InstructGPT3.5 & APE            & Human         & Human + PACE  & StablePrompt(Ours) \\ \midrule
larger animal             & 95.0           & 93.0          & \textbf{95.0} & 93.0              \\
antonyms                   & 80.0           & 85.0          & \textbf{87.0} & 85.0              \\
common concept            & 11.9           & 15.0          & 16.0          & \textbf{24.4}     \\
sentence similarity       & 10.0   & 38.0          & 35.0          & 31.0              \\
synonyms                   & 27.0           & 15.0          & 17.0          & \textbf{43.0}     \\
word in context          & 57.0           & 54.0          & 58.0          & \textbf{60.0}     \\
second letter             & 100.0          & 99.0          & 100.0         & 100.0             \\
cause selection           & 80.0           & 84.0          & 85.0          & \textbf{92.0}     \\
passivization              & 100.0          & 100.0         & 100.0         & 100.0             \\
Translation en-fr          & 87.0           & 89.0          & 88.0          & \textbf{90.0}     \\
sentiment                  & 89.0           & 91.0          & \textbf{92.0} & 90.0              \\
diff                       & 100.0          & 100.0         & 100.0         & 100.0             \\
first word letter        & 100.0          & 100.0         & 100.0         & 100.0             \\
informal to formal       & 50.1           & 64.0          & \textbf{67.0} & 58.0              \\
letters list              & 100.0          & 100.0         & 100.0         & 100.0             \\
negation                   & 76.0           & 79.0          & 83.0          & \textbf{84.0}     \\
num to verbal            & 99.0           & 100.0         & 100.0         & 99.0              \\
ortho starts with        & 68.0           & \textbf{72.0} & 71.0          & 66.0              \\
rhymes                     & \textbf{100.0} & 61.0          & 61.0          & 95.0              \\
singular to plural       & 96.0           & 100.0         & 100.0         & 99.0              \\
sum                        & 100.0          & 100.0         & 100.0         & 100.0             \\
taxonomy animal           & 70.0           & \textbf{98.0} & 96.0          & 75.0              \\
Translation en-es          & \textbf{91.0}  & 90.0          & 89.0          & 89.0              \\
Translation en-de          & 83.0           & \textbf{89.0} & 88.0          & 83.0              \\  \bottomrule
\end{tabular}%
}
\caption{Detail accuracy of 24 tasks of instruction induction datasets with InstructGPT3.5 as target model}
\label{tab:appendix_gpt_full}
\end{table*}
\begin{table*}[]
\resizebox{\textwidth}{!}{%
\begin{tabular}{@{}llllllll@{}}
\toprule
Type &
  Subject &
  \begin{tabular}[c]{@{}l@{}}Fewshot+\\ Manual \\ Prompt\end{tabular} &
  CoT &
  APE &
  ProTeGi &
  StablePrompt &
  \begin{tabular}[c]{@{}l@{}}TTE-\\ StablePrompt\end{tabular} \\ \midrule
\multirow{19}{*}{STEM}       & abstract algebra                    & 30.00 & 33.00 & 31.00 & 35.00 & 32.00 & 33.94 \\
                             & anatomy                             & 50.37 & 51.85 & 49.63 & 52.95 & 54.81 & 56.46 \\
                             & astronomy                           & 57.89 & 64.47 & 53.95 & 56.58 & 64.47 & 60.00 \\
                             & college biology                     & 66.67 & 67.36 & 56.98 & 65.80 & 64.58 & 68.75 \\
                             & college chemistry                   & 38.00 & 34.00 & 39.00 & 40.00 & 43.00 & 39.29 \\
                             & college computer science            & 41.00 & 48.00 & 32.80 & 37.00 & 40.00 & 43.75 \\
                             & college mathematics                 & 32.00 & 34.00 & 33.00 & 33.00 & 34.00 & 40.19 \\
                             & college physics                     & 39.22 & 34.31 & 32.33 & 35.29 & 36.27 & 35.71 \\
                             & computer security                   & 70.00 & 67.00 & 62.20 & 67.00 & 67.00 & 66.07 \\
                             & conceptual physics                  & 51.06 & 55.31 & 51.06 & 49.79 & 49.36 & 49.58 \\
                             & electrical engineering              & 51.72 & 55.17 & 46.21 & 40.00 & 53.10 & 56.34 \\
                             & elementary mathematics              & 38.89 & 60.05 & 38.10 & 37.30 & 39.15 & 44.01 \\
                             & high school biology                 & 70.65 & 64.52 & 65.81 & 69.81 & 71.94 & 70.94 \\
                             & high school chemistry               & 52.71 & 52.71 & 52.22 & 45.82 & 49.26 & 51.44 \\
                             & high school computer science        & 61.00 & 58.00 & 54.00 & 51.00 & 55.00 & 55.00 \\
                             & high school mathematics             & 36.30 & 33.70 & 38.52 & 32.96 & 34.81 & 37.13 \\
                             & high school physics                 & 26.49 & 31.13 & 32.45 & 33.77 & 32.45 & 43.50 \\
                             & high school statistics              & 45.37 & 43.52 & 46.76 & 50.46 & 45.83 & 45.83 \\
                             & machine learning                    & 35.71 & 46.43 & 39.29 & 35.71 & 41.07 & 44.67 \\ \midrule
\multirow{13}{*}{Social Science} &
  econometrics &
  32.46 &
  34.21 &
  32.46 &
  31.58 &
  32.46 &
  40.63 \\
                             & high school geography               & 66.67 & 61.11 & 56.57 & 59.69 & 73.74 & 76.46 \\
                             & high school government and politics & 74.09 & 76.17 & 67.88 & 70.89 & 77.72 & 73.64 \\
                             & high school macroeconomics          & 54.10 & 55.13 & 50.00 & 56.15 & 58.97 & 57.75 \\
                             & high school microeconomics          & 55.46 & 55.46 & 53.36 & 56.15 & 63.03 & 64.58 \\
                             & high school psychology              & 76.33 & 73.58 & 71.19 & 72.66 & 75.78 & 81.64 \\
                             & high school psychology              & 76.33 & 73.58 & 71.19 & 72.66 & 75.78 & 81.64 \\
                             & human sexuality                     & 62.60 & 52.76 & 61.07 & 58.78 & 64.89 & 63.19 \\
                             & professional psychology             & 51.80 & 53.43 & 49.51 & 48.09 & 54.11 & 55.72 \\
                             & public relations                    & 60.00 & 54.55 & 63.64 & 59.09 & 55.67 & 63.39 \\
                             & security studies                    & 50.20 & 48.57 & 52.24 & 47.35 & 50.20 & 50.20 \\
                             & sociology                           & 66.17 & 67.19 & 65.17 & 70.65 & 71.64 & 67.79 \\
                             & us foreign policy                   & 75.00 & 69.00 & 76.00 & 73.00 & 73.00 & 78.00 \\ \midrule
\multirow{13}{*}{Humanities} & formal logic                        & 37.30 & 38.10 & 36.51 & 33.33 & 38.10 & 39.84 \\
                             & high school european history        & 63.64 & 57.58 & 62.42 & 65.45 & 68.48 & 64.84 \\
                             & high school us history              & 62.75 & 56.86 & 65.20 & 55.39 & 70.59 & 70.59 \\
                             & high school world history           & 68.35 & 67.51 & 71.23 & 64.14 & 75.00 & 77.59 \\
                             & international law                   & 61.98 & 65.29 & 64.46 & 66.12 & 71.07 & 67.97 \\
                             & jurisprudence                       & 57.41 & 63.89 & 62.04 & 62.04 & 56.48 & 66.97 \\
                             & logical fallacies                   & 63.19 & 65.03 & 68.10 & 66.87 & 64.42 & 64.74 \\
                             & moral disputes                      & 49.71 & 51.16 & 58.96 & 55.49 & 57.23 & 59.94 \\
                             & moral scenarios                     & 24.36 & 27.93 & 27.26 & 29.27 & 30.50 & 26.67 \\
                             & philosophy                          & 56.91 & 54.66 & 54.66 & 57.23 & 57.23 & 58.75 \\
                             & prehistory                          & 60.49 & 52.16 & 58.64 & 56.17 & 58.95 & 58.96 \\
                             & professional law                    & 40.61 & 38.53 & 32.01 & 41.98 & 38.98 & 43.05 \\
                             & world religions                     & 73.68 & 69.59 & 71.93 & 74.27 & 74.27 & 75.00 \\ \midrule
\multirow{13}{*}{Others}     & business ethics                     & 47.00 & 63.00 & 55.00 & 51.00 & 55.00 & 55.36 \\
                             & clinical knowledge                  & 54.34 & 56.60 & 51.20 & 51.70 & 62.26 & 60.69 \\
                             & college medicine                    & 54.34 & 53.17 & 46.87 & 49.71 & 58.96 & 58.96 \\
                             & global facts                        & 32.00 & 39.00 & 32.00 & 35.00 & 36.00 & 37.50 \\
                             & human aging                         & 56.50 & 55.61 & 58.74 & 58.30 & 58.30 & 61.19 \\
                             & management                          & 61.17 & 64.08 & 63.11 & 60.19 & 68.93 & 74.17 \\
                             & marketing                           & 75.64 & 80.34 & 76.92 & 77.35 & 84.19 & 83.33 \\
                             & medical genetics                    & 54.00 & 55.00 & 55.00 & 57.00 & 56.00 & 58.04 \\
                             & miscellaneous                       & 73.31 & 74.20 & 72.41 & 72.80 & 72.80 & 74.14 \\
                             & nutrition                           & 59.15 & 53.59 & 56.86 & 62.09 & 61.11 & 60.00 \\
                             & professional accounting             & 40.07 & 41.48 & 46.45 & 42.90 & 55.72 & 45.03 \\
                             & professional medicine               & 55.15 & 45.22 & 0.50  & 50.73 & 44.33 & 48.53 \\
                             & virology                            & 46.39 & 47.22 & 48.80 & 50.00 & 53.61 & 47.16 \\ \bottomrule
\end{tabular}%
}
\caption{Full results of MMLU QA datasets.}
\label{tab:appendix_mmlu_full}
\end{table*}


\end{document}